\documentclass[twoside,11pt]{article}

\usepackage{jmlr2e}

\jmlrheading{1}{2023}{}{}{}{H. J. Escalante and A. Kruchinina}

\ShortHeadings{Academic Competitions}{Escalante and Kruchinina}
\firstpageno{1}

\begin{document}

\title{Academic Competitions}

\author{\name Hugo Jair Escalante \email hugo.jair@inaoep.mx \\
       \addr Instituto Nacional de Astrof\'isica,\\ \'Optica y Electr\'onica,\\ 
       Tonantzintla, 72840, Puebla, Mexico 
       \AND
       \name Aleksandra Kruchinina \email aleksandra.kruchinina@universite-paris-saclay.fr \\
       \addr Universit\'e Paris Saclay\\
       Paris, France}

\editor{}

\maketitle

\begin{abstract}
Academic challenges comprise effective means for (i) advancing the state of the art, (ii) putting in the spotlight of a scientific community specific topics and problems, as well as (iii) closing the gap for under represented communities in terms of accessing and participating in the shaping of research fields.   Competitions can be traced back for centuries and their achievements have had great influence in our modern world. Recently,  they (re)gained popularity, with the overwhelming amounts of data that is being generated in different domains, as well as the need of pushing the barriers of existing methods, and available tools to handle such data.
This chapter provides a survey of academic challenges in the context of machine learning and related fields.  We review the most influential competitions in the last few years and analyze  challenges per area of knowledge. The aims of scientific challenges, their goals, major achievements and expectations for the next few years are reviewed.
\end{abstract}


\section{Introduction}
\label{sec:intro}
Competitions are nowadays a key component of academic events, as they comprise effective means for making rapid progress in specific topics. By posing a challenge to the academic community, competition organizers contribute to pushing the state of the art in specific subjects and/or to solve problems of practical importance. In fact, challenges are a channel for the reproducibility and validation of experimental results in specific scenarios and tasks.

\begin{figure}[h]
    \centering
    \includegraphics[scale=0.5]{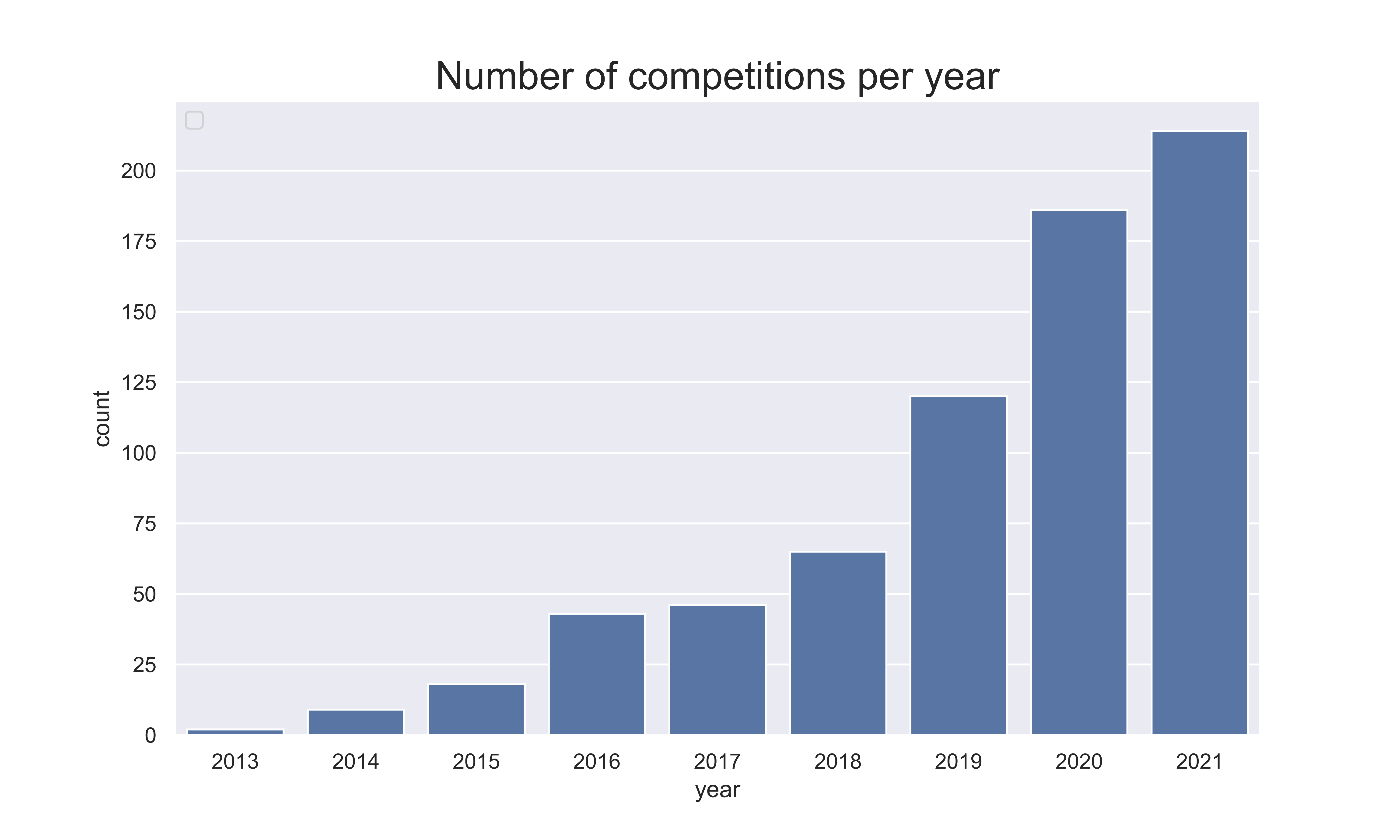}
    \caption{Evolution of the number of competitions each year. Data gathered from {\em CodaLab Competitions} \citep{codalab_competitions_JMLR}, a platform with a community focused on academic competitions. 
    } 
    \label{fig:nbr_compet_years}
\end{figure}

We can distinguish two types of competitions: those associated to industry or aiming at solving a practical problem, and those that are associated to a research question (academic competitions). While sometimes it is  difficult to typecast competitions in these two categories, one can often identify a tendency to either variant. This chapter focuses on academic competitions, although some of the reviewed challenges are often associated to industry too. 
An academic competition can be defined as a \emph{contest that  aims to answer a scientific question via crowd sourcing where participants propose innovative solutions, ideally the challenge will  push the state-of-the-art and have a long-lasting impact and/or an established benchmark.} In this context, academic competitions relying on data have been organized for a while in a number of fields like  natural language processing~\citep{trec93}, machine learning~\citep{NIPS2004_5e751896} and  knowledge discovery in databases\footnote{\url{https://www.kdd.org/kdd-cup/view/kdd-cup-1997}}, however, their spread and impact has 
considerably increased during the last decade, see Figure~\ref{fig:nbr_compet_years} for statistics of the CodaLab platform~\citep{codalab_competitions_JMLR}. 

As a consequence of this growth,  we can witness the permeation and influence that competitions have had in a number of fields.
This chapter aims to survey academic competitions and their impact in the last few years. The objective is to provide the reader with a snapshot of the rise and establishment of academic competitions, and to outline open questions that could be addressed with support of contests in the near future. We have focused on machine learning competitions with emphasis on academic challenges. Nevertheless, competitions from other related fields are also briefly reviewed. 

The remainder of this chapter is organized as follows. Next section provides a brief historical review of competitions in the context of academia and their impact in different fields. 
Then in Section~\ref{sec:perdiscipline:Chapter6} we review academic competitions in terms of the associated field. Finally in Section~\ref{sec:discussion7} we outline some thoughts and ideas on the future of academic competitions. 

\section{A review of academic challenges: past and present}
\label{sec:pastpresent}
This section provides a survey on academic challenges in the context of machine learning and related fields. 

\subsection{Historical review}
\label{sec:historicalreview}



While it is a daunting task to give a comprehensive timeline of the evolution of challenges in machine learning and related fields, this section aims at providing a generic overview. Perhaps the first memorable \emph{challenge} is the Longitude Act issued in 1714. It asked participants to develop a method to determine longitude up to a half degree accuracy (i.e., about 69 miles in distance if one is placed in the Meridian). After years of milestones and fierce competition, Thomas Harrison was acknowledged as the winner of this \emph{challenge}. The main incentive, in addition to scientific curiosity, was a monetary prize offered by the British crown that today would be equivalent to millions of pounds. 

This form of incentive has guided several other competitions organized by governments\footnote{\url{https://www.nasa.gov/solve/history-of-challenges}}, for example 
the DARPA (Defense Advanced Research Projects Agency) grand challenge\footnote{\url{https://www.darpa.mil/news-events/2014-03-13}} series that for years organized competitions for building an all-terrain autonomous vehicle. These type of challenges are still being organized nowadays, not only by governments but also by other institutions and even the private sector. 
Consider for instance the funded challenges organized by the National Institute of Standards and Technology\footnote{\url{https://www.nist.gov/}}  (NIST) and the latest editions of the X-Prize Challenge\footnote{\url{https://www.xprize.org/challenges}} and the  Longitude Prize\footnote{\url{https://longitudeprize.org/}}, both targeting critical  health problems via challenges in their most recent editions. 
This same model of making progress via crowd sourcing has been adopted by academy for a while now. The first efforts in this direction arose in the 90s, it was in that decade that the first RoboCup, ICDAR (International Conference on Document Analysis and Recognition), KDD Cup (Knowledge Discovery and Data Mining Tools Competition) and TREC (Text Retrieval Conference) 
competitions were organized. Such challenges are still being organized on a yearly basis, 
and they have helped to guide the progress in their respective fields. 

RoboCup initially focused on the development of robotic systems able to eventually \emph{play} Soccer at human level~\citep{robocup1}. With currently more than 25 editions, 
RoboCup has evolved in the type of tasks addressed in the context of the challenge. For instance, the 2022 edition\footnote{\url{https://2022.robocup.org/}} comprises leagues on rescue robots, service robots,  soccer playing robots, industrial robots and even a junior league for kids, where each league has multiple tracks. RoboCup competition model has motivated progress on different sub fields within robotics, from hardware to robot control and multi agent communication among others, see~\citep{DBLP:journals/ki/Visser16a}  for a survey on the achievements of this first 20 editions of RoboCup. Together with the DARPA challenge, RoboCup has largely guided the progress of autonomous robotic agents that interact in physical environments.   

Organized by  NIST, TREC is another of the \emph{long-lived} evaluation forums that arose in the early 90s~\citep{trec93}. TREC initially focused on text retrieval tasks. Unlike RoboCup, where solutions were tested lively during the event, TREC asked participants to submit \emph{runs} of their retrieval systems in response to a series of queries. By that time this represented a great opportunity for participants to evaluate their solutions in large scale and realistic retrieval scenarios. This evaluation model actually is still popular among text-based evaluation forums (see e.g., SemEval\footnote{\url{https://semeval.github.io/}}). The TREC forum has evolved and now it focuses on a diversity of tasks around information retrieval (e.g., retrieval of clinical treatments based on patients' cases). Additionally, TREC gave rise to a number of efforts like CLEF (Conference and Labs of the Evaluation Forum), ImageCLEF and TRECVID. They split from TREC to deal with specific sub problems such as: question answering, image and video retrieval, respectively.   


In terms of OCR, there were also efforts aiming to boost research in this open problem during the 90s~\citep{mnistreport97}. The first ICDAR conference took place in 1991, although well documented competitions started in the early 00s 
(see, e.g.,~\citep{DBLP:conf/icdar/LucasPSTWY03}), it seems that competitions associated to digital document analysis were associated to ICDAR since the early 90s, see~\citep{DBLP:conf/icdar/MatsuiNYWY93}.  By that time,  NIST released a large dataset of handwritten digits \citep{Grother1995NISTSD} with detailed instructions on preprocessing, evaluation protocols and reference results. While this was not precisely an academic competition, this effort allowed reproducibility in times where the world was starting to benefit from information spread throughout the internet.  The impact of this effort has been such that, in addition to motivating breakthroughs in OCR, established the MNIST benchmark as a reference problem for supervised learning (see e.g., Yann Lecun's site\footnote{\url{http://yann.lecun.com/exdb/mnist/}} on results in a subset of this benchmark).  Please note MNIST is a \emph{biased} dataset, and other versions of it exist, including QMNIST \citep{qmnist-2019}, where the authors reconstructed the MNIST test set with 60,000 samples. 

Another successful challenge series is the KDD Cup, with its first edition taking place in 1997\footnote{\url{https://www.kdd.org/kdd-cup/view/kdd-cup-1997}}. KDD Cup has focused on challenges on data mining bridging industry and academy, with a variety of topics being covered with time, from retailing, recommendation and customer analysis to authorship analysis and student performance evaluation\footnote{\url{https://kdd.org/kdd-cup}}. While KDD Cup 
has been more application-oriented, findings from this competition have resulted in progress in the field without any doubt. KDD Cups are reviewed in the next chapter.


The first decade of the 2000 was critical for the consolidation of challenges as a way to solve tough problems with the help from the community. It was during this time that the popular Netflix prize\footnote{\url{https://www.netflixprize.com/}} was organized, granting a 1M dollar prize to the team able to improve the performance of their \emph{in-house} recommendation method. The winning team  improved by $\approx 10\%$ the reference model~\citep{bellkor}.  
Also, one of the long-lived competition programs in the context  of machine learning  arose in this decade\footnote{\url{https://sorry.vse.cz/~berka/challenge/PAST/}}: the \emph{ECML/PKDD Discovery Challenge series}. Organized since 1999, this forum has released a number of datasets, although it is now an established competition track, in the early years, competitions consisted of releasing data and asking participants to build and evaluate solutions by themselves. The NeurIPS 2003 feature selection challenge took place\footnote{\url{http://clopinet.com/isabelle/Projects/NIPS2003/}} in this decade too, being this one of the oldest  machine learning competitions in which test data was withheld from participants~\citep{NIPS2004_5e751896}.

In that same decade, the first edition of evaluation efforts that are still being run were launched, for instance, the first: CLEF\footnote{\url{https://www.clef-initiative.eu/web/clef-initiative/}} (2000),   ImageCLEF\footnote{{\url{https://www.imageclef.org/}}} forum (2003),  TRECVID\footnote{\url{https://trecvid.nist.gov/}} conference (2003),  PASCAL VOC\footnote{\url{http://host.robots.ox.ac.uk/pascal/VOC/}} (2005) challenges. All of these efforts and others that evolved over the years (e.g.,  the model selection\footnote{\url{http://clopinet.com/isabelle/Projects/NIPS2006/home.html}} and performance prediction\footnote{\url{http://www.modelselect.inf.ethz.ch/}} challenges (2006) that laid the foundation for AutoML challenges), set the basis for the settlement of academic competitions. 

The 2000s not only were fruitful in terms of the number and variety of longlasting challenges that emerged, but also because of the establishment of organizations. It was in 2009 that Kaggle\footnote{\url{https://kaggle.com/}} was founded, initially focused on challenges as a service, nowadays Kaggle also offers learning, hiring and data-code sharing options. From the academic side, in 2011 ChaLearn\footnote{\url{http://chalearn.org/}}, the Challenges in Machine Learning Organization was founded as well. ChaLearn is a non-profit organization that focuses on the organization and dissemination of academic challenges. ChaLearn provides support to potential organizers of competitions and regularly collaborates with a number of institutions and research groups, likewise, it focuses on research associated to challenge organization in general, this book is a product of such efforts.   

From 2010 and on challenges have been established as one of the most effective way of boosting research in a specific problem to get practical solutions rapidly. 
The ImageNet Large Scale Visual Recognition Challenge (ILSVRC) featured from 2010 to 2017 has been among the most successful challenges in computer vision, as it witnessed the rise of CNNs for solving image classification tasks, see next Section. Likewise, the VOC challenge organized until 2012, contributed to the development of object detection techniques like Yolo~\citep{DBLP:conf/cvpr/RedmonDGF16}.  The AutoML challenge series (from 2015) proved that long term contests with code submission could lead to progress on the automation of model design at different levels. As a result, nowadays, top-conferences and venues from different fields have their competition track. Table~\ref{tab:confchallengetracks} shows representative competition programs  associated to major conferences and related organizations. 
\begin{table}[h]
\centering
\caption{Competition tracks of main conferences in machine learning and related fields. Column four shows the number of tasks organized in the latest edition of the associated track (\# Tasks LE) as of 2022. 
Acronyms are as follows: Machine Learning (ML), Data Mining (DM), Computational Intelligence (CI), Pattern Recognition (PR), Robotics (RO), MIR (Multimedia Information Retrieval), Multimedia Information Processing (MIP), Information Retrieval (IR), Natural Language Processing (NLP), Artificial Intelligence (AI), Evolutionary Computation (EC), Medical Image Analysis (MI), Signal Processing (SP), Image Processing (IP), Miscellaneous (MS). The last four rows of this table shows institutions and organizations associated with challenges.}
\label{tab:confchallengetracks}
\resizebox{11cm}{!}{  
\begin{tabular}{ccccc}
\hline
\textbf{Venue}&\textbf{Field}&\textbf{Since}&\textbf{\# Tasks LE}&\textbf{URL}\\
\hline
TREC&IR&1993&7&\url{https://trec.nist.gov/}\\
ICDAR&PR&1993&13&\url{https://icdar2023.org/}\\
KDD&DM&1997&2&\url{https://kdd.org/}\\
ECML&ML&1999&3&\url{https://ecmlpkdd.org/}\\
RoboCup&RO&1997&5&\url{https://www.robocup.org/}\\
PAN-CLEF$\dag$&NLP&2000&4&\url{https://pan.webis.de/}\\
TrecVid&MIR&2003&8&\url{https://trecvid.nist.gov/}\\
ImageCLEF$\dag$&MIP&2003&4&\url{https://www.imageclef.org}\\
MediaEval&MIP&2003&11&\url{https://multimediaeval.github.io/}\\
GECCO&EC&2004&10&\url{https://gecco-2022.sigevo.org/HomePage}\\
WCCI&CI&2006&13&\url{https://wcci2022.org/accepted-competitions/}\\
MICCAI&MI&2007&38&\url{https://conferences.miccai.org/2022/en/}\\
Interspeech&SP&2008&2&\url{https://interspeech2022.org/}\\
ICRA&RO&2008&10&\url{https://www.icra2022.org/}\\
ACM Multimedia&MIP&2009&10&\url{https://2022.acmmm.org/grand-challenges/}\\
ICPR&PR&2010&7&\url{https://www.icpr2022.com/}\\
SemEval&NLP&2010&12&\url{https://semeval.github.io/}\\
IROS&RO&2012&9&\url{https://iros2022.org/program/competition/}\\
ICMI&MIP&2013&1&\url{https://icmi.acm.org/2022/}\\
ICASSP&SP&2014&8&\url{https://2022.ieeeicassp.org/}\\
ICME&MIP&2015&2&\url{https://2022.ieeeicme.org/}\\
CIKM&DM&2017&2&\url{https://www.cikm2022.org}\\
ICIP&IP&2017&4&\url{https://2022.ieeeicip.org/}\\
NeurIPS&ML&2018&25&\url{https://neurips.cc/Conferences}\\
IJCAI&AI&2018&4&\url{https://www.ijcai.org/}\\
\hline
AutoML&ML&2022&1&\url{https://automl.cc/}\\
\hline

Loingitude Prize&MS&1714$^*$&1&\url{https://longitudeprize.org/}\\
XPrize$^*$&MS&1996&2&\url{https://www.xprize.org/}\\
\hline
Kaggle&MS&2009&-&\url{https://www.kaggle.com/}\\
ChaLearn&ML&2011&-&\url{http://chalearn.org/}\\
\hline
\end{tabular}
}
\end{table}

This table illustrates that many scientific communities have acknowledged the importance of academic competitions, and highly value these by dedicating resources towards organizing such competitions.
Please note that there are top tier venues that do not have an \emph{official} competition track, and therefore they were not included in this table. However, these venues have hosted workshops associated to competitions that have had great impact. Just to name a few: CVPR, ICCV, ECCV, ICML, ICLR, EMNLP, ACL. 

\subsection{Progress driven by academic challenges}
As previously mentioned challenges are now established mechanisms for dealing with complex problems in science and industry. This is not fortuitous, but a response from the community to a number of accomplishments in different fields. This section aims to briefly summarize the main achievements of selected challenges that have motivated other researchers and fields to organize competitions. We focused on a representative machine learning challenge (AutoML) and two evaluation campaigns from the two fields where more contests are organized, see Figure~\ref{fig:pie_chart}.

\begin{itemize}

    \item \textbf{AutoML challenges.} 
    AutoML is the sub field of machine learning that  aims at automating as much as possible all of the aspects of the design cycle~\citep{automlbook}. While people were initially sceptical of the potential of this sort of methods, nowadays AutoML is a trending research topic within machine learning (there is a dedicated  AutoML conference with a competition track\footnote{\url{https://automl.cc/}} since 2022). This is in large part due to the achievements obtained in the context of AutoML challenges. Back in 2006 early efforts in this direction were the prediction performance challenge~\citep{DBLP:conf/ijcnn/GuyonADB06} and the agnostic {\em vs.} prior knowledge challenge~\citep{DBLP:journals/nn/GuyonSDC08}. These contests asked participants to build methods for automatically or manually building classification models. They became the predecessors of the AutoML challenge series that ran from 2015 to 2018~\citep{DBLP:books/sp/19/GuyonSBEELJRSSSTV19},
    and all of the follow up events that are still organized. Initially, the AutoML challenge series focused on tabular data, but it then evolved to deal with raw heterogeneous data in the AutoDL\footnote{ \url{https://autodl.chalearn.org/}} challenge series\citep{ChaLearnAutoDL2019}, whose  latest edition is the  Cross-Domain MetaDL challenge 2022~\footnote{\url{https://metalearning.chalearn.org/}}~\citep{elbaz2021metadl, elbaz:hal-03688638, https://doi.org/10.48550/arxiv.2208.14686}.
    A number of methods (e.g., AutoSKLearn~\citep{Feurer2019}), evaluation protocols, AutoML mechanisms (e.g., Fast Augmentation Learning methods~\citep{baek2020autoclint}) and improvements arose in the context of these challenges including the evaluation of submitted  code, cheating prevention mechanisms, the progressive automation of different types of tasks (e.g., from binary classification to regression, to multiclass classification, to neural architecture search) and the use of different data sources  (from tabular data, to raw images, to raw heterogeneous datasets). The result is an established benchmark that is widely used by the community. 
    
    \item \textbf{ImageNet Large Scale Visual Recognition Challenge.} The so called, ImageNet challenge asked participants to develop image classification systems for 1,000 categories and using millions of images as training data~\citep{Russakovsky2015}. At the time of the first edition of the challenge, object recognition, image retrieval and classification datasets were dealing with problems involving thousands of images and dozens of categories (see e.g.,~\citep{DBLP:journals/cviu/EscalanteHGLMMSPG10}). While the scale made participants struggle in the first two editions of the challenge, the third round witnessed the renaissance of convolutional neural networks, when AlexNet reduced drastically the error rate for this dataset~\citep{DBLP:conf/nips/KrizhevskySH12}. In the following editions of the challenge other landmark CNN-based architectures for image classification were proposed including: VGG~\citep{DBLP:journals/corr/SimonyanZ14a}, GoogLeNet~\citep{DBLP:conf/cvpr/SzegedyLJSRAEVR15} and ResNet~\citep{DBLP:journals/corr/HeZRS15}. These architectures comprised important contributions to deep learning, including residual connections/blocks and inception-based networks, the establishment of regularization mechanisms like dropout, pretraining and fine tuning and the efficient usage of GPUs for training large models. While the challenge itself did not provoke 
    the aforementioned contributions, it was the catalyst and solid 
    test bed for the rise of deep learning in computer vision. 
    
    \item \textbf{Text Retrieval Evaluation Conference.} 
    TREC initially focused on the evaluation of information retrieval systems (text) (see~\citep{DBLP:journals/ipm/Muresan07,OVER2001369} for an overview of the early editions of TREC), but it rapidly evolved to include novel tasks and evaluation scenarios in the forthcoming years. This led 
    to include? tasks that involved information sources from multiple languages, and eventually images and videos. Other tasks that have been widely considered in the TREC campaign are: question answering, adaptive filtering, text summarization, indexing, among many others. Thanks to this effort the information retrieval and text mining fields were consolidated and boosted the progress in the development of search engines and related tools that are quite common nowadays. Well known retrieval models and related mechanisms for efficient indexing, query expansion, relevance feedback, arose in the context of TREC or were validated in this forum. 
    Another  important contribution of TREC through the years is that it has evolved to give rise to numerous tasks and application scenarios that have defined the text mining field. 
    
\end{itemize}

We surveyed a few representative challenges and outlined the main benefits that they bring into their respective communities. While these are very specific examples and while we have chosen breaking through competitions, similar outcomes can be drawn from challenges organized in other fields. In Section~\ref{sec:perdiscipline:Chapter6} we review challenges from a wider variety of domains. 

\subsection{Pros and cons of academic challenges}
We have learned so far that challenges are beneficial in a number of ways, and have boosted progress in a variety of domains. However, it is true that there are some limitations and undesired effects of challenges that deserve to be pointed out. This section briefly elaborates on benefits and limitations of academic challenges. 

\subsubsection{Benefits of academic challenges}
As previously mentioned, the main benefit of challenges is the solution of complex problems via crowd sourcing, advancing the state of the art and the establishment of benchmarks. There are, however, other benefits that make them appealing to both participants and organizers, these include:   
\begin{itemize}
    \item \textbf{Training and learning through challenges.} Competitions are an effective way to learn new skills, they \emph{challenge} participants to gain new knowledge and put in practice known concepts for solving relevant problems in research and industry. Even if participants do not win a challenge or a series of them, they progressively improve their problem solving skills.  
    \item \textbf{Challenges are open to anyone.} Apart of political restrictions that may be applied for some organizations, competitions target anyone with the ability to approach the posted problem. This is particularly appealing to underrepresented groups and people with limitations to access the cutting edge problems, data and resources. For instance, most competitions adopting code submission provide cloud-based computing to participants. 
    Likewise, challenges can be turned into ever lasting benchmarks and they contribute to making data available to the public. 
    \item \textbf{Engagement and motivation.} The engagement offered by competitions is priceless. Whether the reward is economic, academic (e.g., publication or talk in a workshop, professional recognition in the field), competitiveness, or just fun, participants find challenges motivating. 
    \item \textbf{Reproducibility.} This cannot be emphasized enough,  benchmarks associated to challenges not only provide the task, data and evaluation protocols. In most cases resources, starting-kits, others' participants code and computing resources are given as well. This represents an easy way to get into competitions to participants, which can directly compete with state-of-the-art solutions. At the same time, competitions having these features 
    guarantee reproducibility of results which is clearly beneficial to the progress in the field. 
\end{itemize} 

\subsubsection{Pitfalls of academic challenges}
Despite the benefits of challenges, they are not risk-free, therefore, there are certain limitations that should be taken into account. 
\begin{itemize}
    \item \textbf{Performance improvement vs. scientific contribution.} Academic challenges often ask participants to build solutions that achieve the best performance according to a given metric. Although in most cases there is a research question associated to a challenge, participants may end up building solutions that optimize the metric but that do not necessarily result in new knowledge. This gives challenges a bitter-sweet taste, as often new findings are overshadow by super-tuned off-the-shell solutions. 
    \item \textbf{Stagnation.} An undesirable outcome for a challenge is stagnation, this is often the result of wrong challenge design decisions, that result in either a problem that is too hard to be solved with current technology or unattractive to participants. While it is not possible to anticipate how far the community can go in solving a task, the implementation of (strong) baselines, starting kits and appealing datasets, or rewards could help to avoid stagnation.    
    \item \textbf{Data Leakage.} It refers to the use of target (or any other relevant information that is supposed to be withheld from participants) information by participants to build their solutions~\citep{LeakageSIGKDD}. This is a common issue when datasets are re-used or when datasets are build from external information (e.g., from social networks). Anonymization and other mechanisms as those exposed in~\citep{LeakageSIGKDD} could be adopted for avoiding this problem.    
    \item \textbf{Privacy and rights on data.} \emph{''Data is the new oil"} has been a popular say recently\footnote{\url{https://www.forbes.com/sites/forbestechcouncil/2019/11/15/data-is-the-new-oil-and-thats-a-good-thing/?sh=381ec30d7304}}, while this is debatable, it is true that data is a valuable asset that must be \emph{handled with care}. Therefore copyright infringement should be avoided to the uttermost end. Likewise, failing to guarantee  privacy is an important issue that must be addressed by organizers as this could lead into legal issues. Anonymization mechanism should be applied to data before its release, making sure it is not possible to track users identity or other important and confidential information.   
    
\end{itemize}

\subsection{What makes academic challenges successful?}
\label{sec:successful_challenges}
Having reviewed competitions, their benefits and pitfalls/limitations, this section elaborates on characteristics that we think make a challenge successful. While it is subjective to define a successful challenge, the following guidelines 
associate success to high participation, quantitative performance and novelty of top ranked solutions.
\begin{itemize}
    \item \textbf{Scientific rigour.} The design and the analysis of the outcomes of a competition are critical for its success. Following  scientific rigor as ``to ensure robust and unbiased experimental design, methodology, analysis, interpretation and reporting of results``~\citep{10.1093/carcin/bgx085} is necessary and helps to avoid some of the  limitations mentioned above. Adopting statistical testing for the analysis of results, careful designing of evaluation metrics, establishing theoretical bounds on these,  running multiple tests before releasing the data/competition,  formalizing the problem formulation, performing ablation studies are all critical actions that impact on the outcomes of academic challenges. 
    
    \item \textbf{Rewarding and praising scientific merit and novelty of solutions.} It is worth mentioning that novel methods do not always make it to the top of the leaderboard, but these new ideas may be great seeds and serve as an inspiration to others for further fruitful research. Therefore, rewarding and acknowledging scientific merit and novelty of solutions is very important. There are several ways of doing this, for instance, having a \emph{prize} for the most original/novel submission or granting a best paper award that is not entirely based on quantitative performance.

    \item \textbf{Publication and dissemination of results} are  good practices with multiple benefits. Participants are often invited to fill out  \emph{fact sheets} and write workshop papers in order to document their solutions. Similarly, organizers  commonly publish overview papers that summarize the competition, highlighting the main findings and analyzing results in detail.  Associating a special issue of a journal 
    with competitions is a good idea as it is motivating for participants, and at the same time it is a \emph{product} that organizers can report in their work evaluations. 
    
    \item \textbf{Associating the competition with an top tier venue} (e.g., conferences, summits, workshops, etc.) makes a challenge more attractive to participants,  as they associate the quality of associated venues and competitions. Also, physically attending the competition session is more appealing if participants can also attend top tier events. 
    
    \item \textbf{Organization of panels and informal discussion sessions} involving both participants and organizers is valuable for sensing perception of people associated to the event. This is critical when organizing challenges that run for several editions. 
    
    \item \textbf{Establishing benchmarks} should be an underlying goal of every competition. Therefore, curated data, fail safe evaluation protocols, and adequate platforms for maintaining competitions as long term evaluation test beds are essential. Likewise,  the use of open data and open source code for the purposes of  reproducibility and so that everyone can benefit and continue their own research.

\end{itemize}

\subsubsection{Academic vs. industrial challenges}
Industrial challenges are described in detail in the next chapter. In this section we  outline the main  differences of industry and scientific competitions. 

The main objective of industrial challenges is the economic advantage from the winning model that will potentially increase profits and improve business model, meaning it should be an end-to-end solution. 

The organizers care much less about scientific publications, being scientifically rigorous neither about the results being statistically significant. 
These types of contest do not single out scientific questions, that is not the priority for them. They aim at specific business problems, usually posses big not preprocessed datasets and evidently can provide more often big prizes. 
Up till now the direct positive correlation between these big rewards and qualitative contributions has not been proven. But it was observed that big prizes might attract many participants, create "big splash" in the news for the company-organizer and cause a lot of noise in the leaderboard, potentially leading to gaining by chance.
While the winners and contributors of academic challenges get scientific recognition, the top performers at the industrial contests can receive job offers and be hired by the organizers. 

Another important aspect of industrial challenges is that due to their nature and the concurrent market, the company-organizers prefer to keep the data and the submitted code private, which is in the opposition with the scientific mentality, because it prevents to benefit from the latest break-through and get inspiration from the newest ideas.



\section{Academic challenges across different fields}
\label{sec:perdiscipline:Chapter6}
\begin{figure}[h]
    \centering
    \includegraphics[scale=0.25]{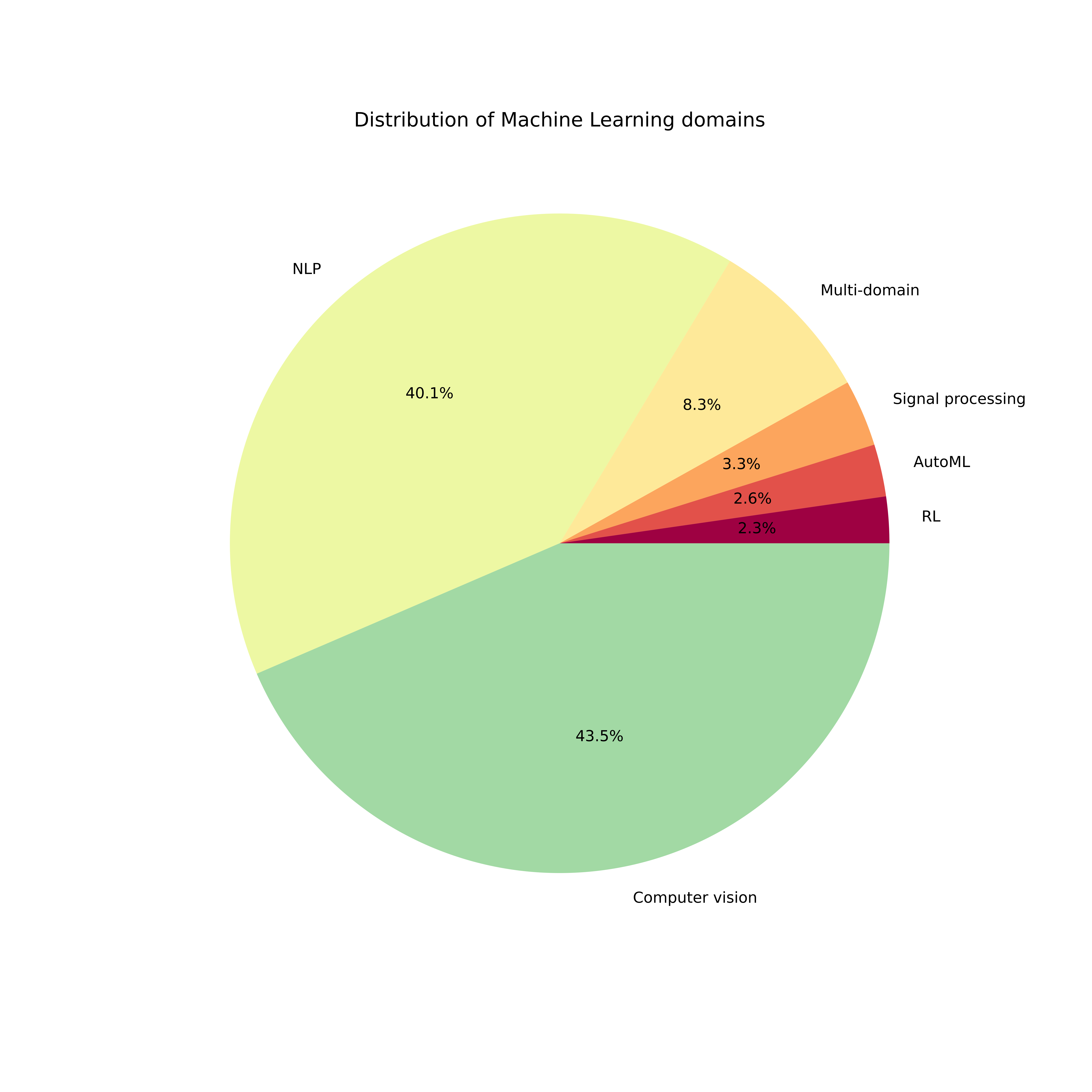}
    \caption{Distribution of competitions with different machine learning domains. Data gathered from CodaLab Competitions~\citep{codalab_competitions_JMLR}}
    \label{fig:pie_chart}
\end{figure}

This section briefly reviews challenges across different fields. We focus on fields that have long tradition in challenges. In order to identify such fields of knowledge, we surveyed competitions organized in the CodaLab platform~\citep{codalab_competitions_JMLR}. Figure~\ref{fig:pie_chart} shows a distribution of CodaLab challenges across fields of knowledge. Clearly NLP and Computer vision challenges dominate, this could be due to the explosion of availability of visual and textual data of the last few years. One should note that most of the competitions shown in that plot have a strong machine learning component. In the remainder of this section we briefly survey competitions organized in a subset of selected fields. 

\subsection{Challenges in Machine Learning}

Machine learning is a transversal field of knowledge that has been present in most challenges regardless of the application field (e.g., computer vision, OCR, NLP, time series analysis, and so on). Therefore, it is not easy to cast a challenge as a ML competition. For that reason, in this section we review as a representative sample the competition track of the NeurIPS conference. The track has run regularly since 2017, although challenges organized with the conference date back to the early 2000s~\citep{NIPS2004_5e751896}. Overview papers for the NeurIPS competition track from 2019 to 2021 can be found in~\citep{DBLP:conf/nips/EscalanteH19,DBLP:conf/nips/EscalanteH20,pmlr-v176-kiela22a}. 

\begin{figure}[h!tb]
    \centering
    \includegraphics[scale=0.4]{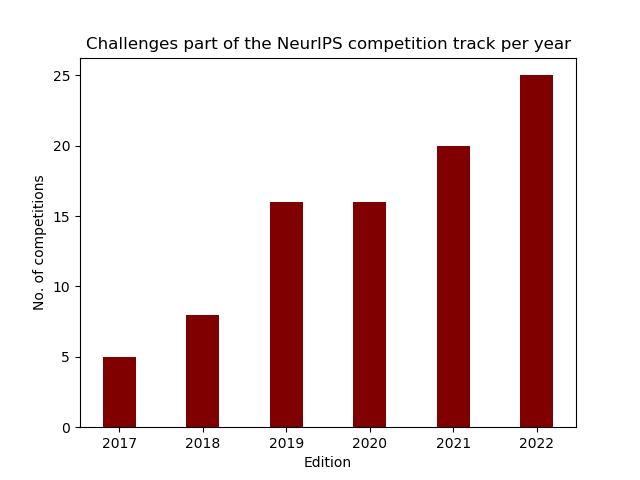}
    \caption{Number of challenges organized as  part of the NeurIPS competition program.
    } 
    \label{fig:neuripsc}
\end{figure}

Figure~\ref{fig:neuripsc} shows the number of competitions that have been part of the  NeurIPS competition track. There has been an increasing number of competitions organized each year, see also~\citep{carlens2023state} for more details. 
The topics of challenges are quite diverse,  with deep reinforcement learning (DRL)  prevailing since the very beginning of the track. The first competition in the program around this topic was the Learning to Run challenge\footnote{\url{https://www.aicrowd.com/challenges/nips-2017-learning-to-run}} that asked participants to build an human-like agent to navigate an environment with obstacles~\citep{DBLP:journals/corr/abs-1804-00361}, this challenge was run for two more editions, the last one being the Learn to Move - Walk Around\footnote{\url{https://www.aicrowd.com/challenges/neurips-2019-learn-to-move-walk-around}} challenge. DRL-based competitions addressing other challenging navigation scenarios are the Animal Olympics\footnote{http://animalaiolympics.com/AAI/} and MineRL series, see below. DRL challenges addressing different tasks are 
the Real robot challenge\footnote{\url{https://real-robot-challenge.com/}} series with two editions, the Learning to run a power network competition\footnote{\url{https://l2rpn.chalearn.org/}} and the two editions of the Pommerman\footnote{\url{https://www.pommerman.com/}} competition where the goal was to develop agents to compete to each other in a bomberman-game-like scenario.  The presence of DRL in the challenge track as been growing in the last editions. 

Another popular topic in the NeurIPS competition track is AutoML: since 2018, at least one competition associated to this topic has been part of  the NeurIPS competition track. These include the AutoML@NeurIPS~\citep{DBLP:journals/corr/abs-1903-05263}  and AutoDL~\citep{ChaLearnAutoDL2019} challenges, the black-box optimization competition~\citep{pmlr-v133-turner21a}, the predicting generalization in deep learning challenge~\footnote{\url{https://sites.google.com/view/pgdl2020}}, two editions of the Meta-DL challenge~\citep{elbaz:hal-03688638,https://doi.org/10.48550/arxiv.2208.14686} and the AutoML Decathlon\footnote{\url{https://www.cs.cmu.edu/~automl-decathlon-22/}}. 

Specific  challenges that have been part of the competition track for more than 2 editions are the following: 
\begin{itemize}
    \item \textbf{Traffic4cast}\footnote{\url{https://www.iarai.ac.at/traffic4cast/}}. Organizing variants of challenges aiming to predict traffic conditions under different settings and scenarios, see~\citep{pmlr-v123-kreil20a,pmlr-v133-kopp21a,pmlr-v176-eichenberger22a}. 
    \item \textbf{The AI Driving Olympics (AI-DO).} Aiming to build autonomous driving systems running in simulation and small physical vehicles tested live during the competition track\footnote{\url{https://www.duckietown.org/research/AI-Driving-olympics}}.
    \item \textbf{MineRL\footnote{\url{https://minerl.io/}}} A competition series focusing on building autonomous agents that using minimal resources are able to solve very complex tasks in a MineCraft environment. In the first two editions agents were asked to find a diamond  with limited resources, see~\citep{pmlr-v123-milani20a,pmlr-v133-guss21a}. In the most recent editions tasks have been varied and more specific~\citep{https://doi.org/10.48550/arxiv.2204.07123}.  
    \item \textbf{Reconnaissance Blind Chess.} Challenges participants to build agents able to play a chess variant in which a player cannot see her
opponent’s pieces but can learn about them through private, explicit sensing actions. Three editions of this competition have run in the track~\citep{pmlr-v123-gardner20a}. 
\end{itemize}

It is difficult to summarize the number and variety of topics addressed in challenges part of the NeurIPS competition, however, we have reviewed a representative sample. Nevertheless, please note that most challenges reviewed in the remainder of this section also  include an ML component. 

\subsection{Challenges in Computer Vision}
Together with machine learning, computer vision has been greatly benefited from challenges. As previously mentioned, The PASCAL Object detection challenge series boosted research on object detection and semantic segmentation~\citep{Everingham15}. The ImageNet large scale classification challenge is another landmark competition that served as platform for the renaissance of convolutional neural networks~\citep{Russakovsky2015}. In addition to these landmark competitions there have been a number of  efforts that have pushed further the state-of-the-art, these are reviewed in the following lines.

The ChaLearn Looking at People (ChaLearn LAP\footnote{\url{https://chalearnlap.cvc.uab.cat/}}) series has organized academic challenges around the  analysis of human behavior from visual information. More than 20 competitions on the topic have been organized so far, see~\citep{DBLP:conf/ijcnn/EscaleraBEG17} for a (outdated) review.  Among the organized competitions several of the datasets have become a reference for different tasks, and are used as benchmarks. These include: the gesture recognition challenges~\citep{DBLP:conf/icmi/EscaleraGBRLGAE13,DBLP:conf/eccv/EscaleraBGBMRPE14, DBLP:books/sp/EGA2017,DBLP:conf/iccvw/WanEAEBGMAGLX17}, the personality recognition challenge series~\citep{DBLP:conf/ijcnn/EscalanteGEJMBA17,DBLP:journals/taffco/EscalanteKSEGGB22,DBLP:conf/iccv/PalmeroBJCNCSSZ21}, the age estimation challenge series~\citep{DBLP:conf/iccvw/EscaleraFPBGEMS15,DBLP:conf/cvpr/EscaleraTMBEGTC16} and the face anti-spoofing challenge series~\citep{DBLP:conf/cvpr/Liu0EETYWLGGL19,DBLP:series/synthesis/2020Wan,DBLP:journals/iet-bmt/LiuL0LEEM0WYTYY21}. A wide diversity of related topics have been studied in the context of ChaLearn LAP challenges, including: action recognition and cultural event recognition~\citep{DBLP:conf/cvpr/BaroGFBOEGE15,DBLP:conf/iccvw/EscaleraFPBGEMS15}, sign language understanding~\citep{DBLP:conf/cvpr/SincanJEK21}, identity preserving human analysis~\citep{DBLP:conf/fgr/ClapesJME20} among others. Undoubtedly, these challenges have advanced the state of the art in a number of directions within computer vision and affective computing. 

The Common Objects in COntext  (COCO\footnote{\url{https://cocodataset.org}})  challenge series that emerged after the end of the Pascal VOC challenge. This effort continued benchmarking object detection methods, but also started evaluating the so called \emph{image captioning} task. Early efforts for the evaluation of this task emerged in the ImageCLEF forum~\citep{DBLP:books/daglib/p/CloughMS10,DBLP:journals/cviu/EscalanteHGLMMSPG10}, where the goal was associating keywords to images. The COCO challenge was more ambitious by asking participants to describe the content of an image with a more \emph{human-like} description. Running from 2015-2020 this benchmark was critical for the consolidation of the image captioning task, with major contributions being reported at the beginning of the series, see~\citep{DBLP:journals/ijon/BaiA18,DBLP:journals/corr/abs-2107-06912}. Today, COCO is an established benchmark in a number of tasks related to vision and language, see~\citep{DBLP:conf/eccv/LinMBHPRDZ14}. 

Other  efforts in the field of computer vision are the NTIRE challenge, focused on image restoration, super resolution and enhancement~\citep{DBLP:conf/cvpr/TimofteAG0ZLSKN17,ntire22} , the visual question answering competition\footnote{\url{https://visualqa.org/}} running from 2016 to 2021, the fine grained classification workshop~\footnote{\url{https://sites.google.com/view/fgvc9}} that has  run a competition program since 2017, the EmotioNet\footnote{\url{https://cbcsl.ece.ohio-state.edu/enc-2020/index.html}} recognition challenge that ran in 2020 and is now a testbed for emotion recognition, the ActivityNet  challenge~\footnote{\url{http://activity-net.org/challenges/2022/}} organized since 2016 and  targeting action recognition  in video, among several others. 


\subsection{Challenges in Natural Language Processing}
The development of the natural language processing (NLP) field, in particular for text mining and related tasks, has been largely driven by competitions, also known in the NLP jargon as \emph{shared tasks}. In fact, one of the oldest evaluation forums across all computer science is one focusing in NLP, that is TREC.  
It initially focused on the evaluation of information retrieval systems (text), but it rapidly evolved to include novel tasks and evaluation scenarios in the forthcoming years~\citep{DBLP:conf/tipster/VoorheesH98,DBLP:journals/ipm/Muresan07,OVER2001369}. This lead to consider tasks that involved information sources from multiple languages~\citep{harman:trec}, and eventually, speech signals~\citep{10.5555/2835865.2835867}  and visual information~\citep{DBLP:journals/corr/abs-2104-13473}. Other tasks that have been  considered in the TREC campaign are: question answering~\citep{DBLP:journals/nle/Voorhees01}, adaptive filtering~\citep{DBLP:conf/trec/Harman95}, text summarization~\footnote{\url{http://trecrts.github.io/}}, among many others. Thanks to this effort the information retrieval and text mining fields were consolidated and boosted the progress in the development of search engines and related tools that are quite common nowadays. 

Several well known evaluation campaigns evolved from TREC and consolidated on their own. Most notably, the TRECVid~\citep{DBLP:journals/corr/abs-2104-13473} and Cross-Language Evaluation Forum~\citep{10.1007/3-540-44645-1_9} (CLEF) campaigns. The former focusing on tasks related to video retrieval, indexing and analysis. The academic and economic impact of TRECVid has been summarized already. Showing the relevance that such forum has had into the progress of video search technology. CLEF is another forum that initially focused on cross-lingual text analysis tasks. Now it is a conference that comprises several shared tasks, called labs. This include ImageCLEF, PAN among others. 
Likewise, there are forums dedicated to specific languages, for example, Evalita\footnote{\url{https://www.evalita.it/campaigns/evalita-2022/}} (for Italian), IberLEF\footnote{\url{https://sites.google.com/view/iberlef2022}} (for Spanish) and GermEval\footnote{\url{https://germeval.github.io/}}. 

In terms of speech, there were several efforts from DARPA~\citep{marcus-1992-overview,black-eskenazi-2009-spoken} and NIST\footnote{\url{https://www.nist.gov/itl/iad/mig/past-hlt-evaluation-projects}} in organizing competitions as early as the late 80s. These long term efforts have helped to shape ASR and related fields. 
More recently, after the deep learning empowering, several challenges focusing on speech have been proposed, these are often associated to major conferences in the field (e.g. Interspeech and ICASSP), see Table~\ref{tab:confchallengetracks}. 
There is no doubt that competitions have played a key role for the shaping the wide field of NLP.

\subsection{Challenges in Biology}

Biology is a field of knowledge that has benefited from competitions considerably. In terms of medical imaging, the premier forum is the  grand challenge series associated to the MICCAI conference, running since 2007~\footnote{\url{https://cause07.grand-challenge.org/Results/}}. A number of important challenges have been organized in this context, where most competitions deal with  medical imagery segmentation or reconstruction of different organs, body parts and input type, see e.g.,~\citep{scully,Marak_Cousty_Najman_Talbot2009,10.1007/978-3-030-98253-9_1}. In recent editions the challenge scenarios and approached tasks have been increasing difficulty  and the potential impact of solutions. 
In its last edition, the MICCAI grand challenge series has 38 competitions running in parallel. This is an indicator of success among the medical imaging community.

Other challenges associated to medical image analysis have been presented in forums associated to image processing and computer vision as well. For instance, in 2019 during The IEEE International Symposium on Biomedical Imaging (ISBI), 
nine challenges were organized\footnote{\url{https://biomedicalimaging.org/2019/challenges/}}. 
In 2020 a challenge on Image processing on real-time distortion classification in laparoscopic videos was organized with 
ICIP 2020\footnote{\url{https://2020.ieeeicip.org/challenge/real-time-distortion-classification-in-laparoscopic-videos/}}. 
In the context of ICCV, 
challenges on remote measurement of physiological signals from videos (RePSS) were organized: one on measurement of inter-beat-intervals (IBI) from facial videos, and another one on respiration measurement from facial videos~\citep{repss1, repss2}.

It is worth mentioning that there are  platforms associated with challenges in biology and medical sciences. The Grand Challenge\footnote{\url{https://grand-challenge.org/challenges/}} platform being perhaps the oldest one and the most representative in terms of imagery: \emph{more than 150 competitions are listed in the platform}, most of which are associated to medical image analysis. A related effort is that of the DREAM challenges\footnote{\url{https://dreamchallenges.org/closed-challenges/}} a platform that has organized more than 60 challenges in biology and medicine. The variety of topics covered by DREAM challenges is vast~\citep{https://doi.org/10.1111/j.1749-6632.2009.04497.x}: from systems biology modelling~\citep{MEYER2021636}, to prevention~\citep{Tarca2020.06.05.130971} and monitoring~\citep{10.1001/jamanetworkopen.2022.27423} damage caused by certain conditions, to disease susceptibility\footnote{\url{https://dreamchallenges.org/respiratory-viral-dream-challenge/}}, to analyzing medical documents with NLP\footnote{\url{https://dreamchallenges.org/electronic-medical-record-nlp-dream-challenge/}}, to drug analysis and combination\footnote{\url{https://dreamchallenges.org/astrazeneca-sanger-drug-combination-prediction-dream-challenge/}} and many other relevant topics. As seen in Chapter 5, platforms play a key role in challenge success, biology is a  field where excellent platforms are available and this has been critical for the advancement of state of the art in this relevant field.

Protein structure modelling was officially introduced in 1994 at the biennial large-scale experiment Critical Assessment of protein Structure Prediction (CASP), and ever since it attracted more than 100 teams to tackle the problem, see~\citep{CASP04}. Only almost 20 years later, two teams presented breaking through solutions to protein folding task~\citep{CASP21}: DeepMind with their AlphaFold2~\citep{Jumper2021HighlyAP} and scientists of the University of Washington with RoseTTAFold~\citep{RosettaFold}. 
Alphafold uses multiple neural networks that feed into each other in two stages. It starts with a network that reads and folds the amino acid sequence and adjusts how far apart pairs of amino acids are in the overall structure. Then goes the structure model network that reads the produced data, creates a 3D structure, and makes the needed adjustments~\citep{casp13,Jumper2021HighlyAP}.
RoseTTAFold adds a simultaneous third neural network, which tracks where the amino acids are in 3D space as the structure folds, alongside the 1D and 2D information~\citep{RosettaFold}. The solution of Washington University is less accurate but uses less computational and time resources than AlphaFold2. Without the existence of the CASP experiment, achieving the outstanding performance of these methods would have taken much more time. 

As we can see, advancements of machine learning in biology are of crucial importance, that’s why there are numerous competitions in this domain. Researchers and practitioners are trying to deal with biological and related domain (medicine, agriculture, and others) challenges using various machine learning solutions like computer vision, NLP and signal processing. 

\subsection{Challenges in Autonomous Driving}
DARPA Grand Challenge is considered as one of the first long distance race for autonomous driving cars, it was organised in 2004 with more than 100 teams. None of the robot vehicles managed to finish the 240 km route, only one member covered 11.78 km and then got stuck. Next year there were 195 teams, the distance of the challenge was of 212 km, and five vehicles successfully completed the course. These first courses were challenging but vehicles “operated in isolation”, their interaction was not required, and there was no traffic either. So the next Urban challenge was held in 2007 in a city area, the objective was to complete 96km in 6 hours and it included “driving on roads, handling intersections and maneuvering in zones”~\citep{DarpaTartanTeam}. Six teams managed to complete the course. 

The basics were laid, and DARPA pursued their competitions: Robotics Challenge in 2012, 2013 - Fast Adaptable Next-Generation Ground Vehicle Challenge, 2013 – 2017 Subterranean Challenge on “autonomous systems to map, navigate, and search underground tunnel, urban, and cave spaces” \footnote{https://www.darpa.mil/about-us/subterranean-challenge-final-event}.

Being able to test autonomous driving cars "in the wild" is important and expensive. In order to fine-grain the algorithms at a less cost one needs to test them virtually. Hopefully, there are different simulators: CARLA \footnote{https://carla.org/}, VISTA 2.0 \footnote{https://vista.csail.mit.edu}, NVIDIA DRIVE Sim \footnote{https://www.nvidia.com/en-us/self-driving-cars/simulation/} and others.

Several challenges have been organised based on  \href{https://carla.org/}{CARLA simulator}, "an open-source simulator for autonomous driving research", which is used to study "a classic modular pipeline, a deep network trained end-to-end via
imitation learning, and a deep network trained via reinforcement learning" \citep{Dosovitskiy17}.

Autonomous driving has numerous interesting challenges, and object detection is one of them. Most of the current research concentrates around camera images, but it is not the best sensor under certain conditions like bad weather, poor lighting. Radar information can help to overcome these inconveniences. It is more reliable, cost-efficient and might potentially lead to better object detection. ROD2021 Challenge is the first competition of its' kind, which proposes object detection task on radar data, and was held in the ACM International Conference on Multimedia Retrieval (ICMR) 2021. Organisers developed their own baseline: “radar object detection pipeline, which consists of two parts: a teacher and a student. Teacher’s pipeline fuses the results from both RGB and RF images
to obtain the object classes and locations in RF images. Student’s pipeline
utilizes only RF images as the input to predict the corresponding ConfMaps
under the teacher’s supervision. The LNMS as post-processing is followed
to calculate the final radar object detection results.” \citep{9353210}.


This challenge attracted more than 260 participants among 37 teams with around 700 submissions. The winning team, affiliated to Baidu, submitted paper “DANet: Dimension Apart Network for Radar Object Detection” \citep{DANet}, where they presented their results. "This paper proposes a dimension apart network (DANet), including a lightweight dimension apart module (DAM) for temporal-spatial feature extraction. The proposed DAM extracts features from each dimension separately and then concatenates the features together. This module has much smaller number of parameters, compared with RODNet-HGwI, so that significant reduction of the computational cost can be achieved. Besides, a vast amount of data augmentations are used for the network training, e.g., mirror, resize, random combination, Gaussian noise and reverse temporal sequence. Finally, an ensemble technique is implemented with a scene classification for a more robust model. The DANet achieves the first place in the ROD2021 Challenge. This method has relatively high performance but with less computational cost, which is an impressive network model. Besides, this method shows data augmentation and ensemble techniques can greatly boost the performance of the radar object detection results" \citep{wang2021rod2021}.
 
Another interesting and pioneering challenge is OmniCV (Omnidirectional Computer Vision) in conjunction with IEEE Computer Society Conference on Computer Vision and Pattern Recognition (CVPR'2021). The objective was to evaluate semantic segmentation techniques targeted for fisheye camera perception. It attracted 71 teams and a total of 395 submissions. Organisers proposed their baseline “a PSPNet network with a ResNet50 backbone finetuned on WoodScape Dataset”, which "achieved a score of 0.56 (mIoU 0.50, accuracy 0.67) excluding void class". The top teams managed to get significantly better scores and proposed interesting solutions. The winning team implemented full Swin-transformer Encoder-Decoder approach, with a score of 0.84 (mIoU 0.86, accuracy 0.89) \citep{Fisheye_AD}.
















\section{Discussion}
\label{sec:discussion7}
Academic challenges have been decisive for the consolidation of fields of knowledge. This chapter provided an historical review and an analysis of benefits and limitations of challenges, while it is true that competitions can have undesired effects, there is palpable evidence that they have boosted research across a number of fields. In fact there are several examples of breakthrough discoveries that have arosen in the context of academic competitions.  

While we are witnessing the establishment of academic competitions as a way to advance the state of the art, the forthcoming years are promising. Specifically, we consider that the following lines of research will be decisive in the next few years:
\begin{itemize}
    \item \textbf{Data centric competitions\footnote{\url{https://https-deeplearning-ai.github.io/data-centric-comp}}} This is competitions where the goal is to improve a dataset by applying so called, data-centric techniques, like fixing mislabeled samples, finding prototypes, border points, summarization, data augmentation, etc. 
    \item \textbf{Cooperative competitions.} Coopetitions is a form of crowd sourcing in which participants compete to build the best solution for a problem, but they cooperate with other participants in order to obtain an additional reward (e.g., information from other participants, higher scores, etc.).
    \item \textbf{Challenges for education.} Exploiting the full potential of challenges in education is a challenge itself, but we think this is  a valuable resource for reaching wider audiences with assignments that require solving practical problems.
    \item \textbf{Academic challenges for good.} This is a topic being pursued and encouraged by evaluation forums and competition tracks, consider for instance the NeurIPS competition track~\citep{DBLP:conf/nips/EscalanteH19,DBLP:conf/nips/EscalanteH20,pmlr-v176-kiela22a}. 
    \item \textbf{Dedicated publications for challenges.} There are few dedicated forums in which results of challenges are published (consider for instance the Challenges in Machine Learning series\footnote{\url{https://www.springer.com/series/15602}}). We foresee more dedicated venues will be available in the next few years.   
\end{itemize}



\acks{We would like to acknowledge support for this project
from the National Science Foundation (NSF grant IIS-9988642)
and the Multidisciplinary Research Program of the Department
of Defense (MURI N00014-00-1-0637). }










\vskip 0.2in
\bibliography{ref}

\begin{thebibliography}{98}
\providecommand{\natexlab}[1]{#1}
\providecommand{\url}[1]{\texttt{#1}}
\expandafter\ifx\csname urlstyle\endcsname\relax
  \providecommand{\doi}[1]{doi: #1}\else
  \providecommand{\doi}{doi: \begingroup \urlstyle{rm}\Url}\fi

\bibitem[Andrearczyk et~al.(2022)Andrearczyk, Oreiller, Boughdad, Rest,
  Elhalawani, Jreige, Prior, Valli{\`e}res, Visvikis, Hatt, and
  Depeursinge]{10.1007/978-3-030-98253-9_1}
Vincent Andrearczyk, Valentin Oreiller, Sarah Boughdad, Catherine Cheze~Le
  Rest, Hesham Elhalawani, Mario Jreige, John~O. Prior, Martin Valli{\`e}res,
  Dimitris Visvikis, Mathieu Hatt, and Adrien Depeursinge.
\newblock Overview of the hecktor challenge at miccai 2021: Automatic head and
  neck tumor segmentation and outcome prediction in pet/ct images.
\newblock In Vincent Andrearczyk, Valentin Oreiller, Mathieu Hatt, and Adrien
  Depeursinge, editors, \emph{Head and Neck Tumor Segmentation and Outcome
  Prediction}, pages 1--37, Cham, 2022. Springer International Publishing.
\newblock ISBN 978-3-030-98253-9.

\bibitem[Awad et~al.(2021)Awad, Butt, Curtis, Fiscus, Godil, Lee, Delgado,
  Zhang, Godard, Chocot, Diduch, Liu, Smeaton, Graham, Jones, Kraaij, and
  Qu{\'{e}}not]{DBLP:journals/corr/abs-2104-13473}
George Awad, Asad~A. Butt, Keith Curtis, Jonathan~G. Fiscus, Afzal Godil,
  Yooyoung Lee, Andrew Delgado, Jesse Zhang, Eliot Godard, Baptiste Chocot,
  Lukas~L. Diduch, Jeffrey Liu, Alan~F. Smeaton, Yvette Graham, Gareth J.~F.
  Jones, Wessel Kraaij, and Georges Qu{\'{e}}not.
\newblock {TRECVID} 2020: {A} comprehensive campaign for evaluating video
  retrieval tasks across multiple application domains.
\newblock \emph{CoRR}, abs/2104.13473, 2021.
\newblock URL \url{https://arxiv.org/abs/2104.13473}.

\bibitem[Baek et~al.(2021)Baek, DiMaio, Anishchenko, Dauparas, Ovchinnikov,
  Lee, Wang, Cong, Kinch, Schaeffer, Millán, Park, Adams, Glassman,
  Degiovanni, Pereira, Rodrigues, Dijk, Ebrecht, and Baker]{RosettaFold}
Minkyung Baek, Frank DiMaio, Ivan Anishchenko, Justas Dauparas, Sergey
  Ovchinnikov, Gyu Lee, Jue Wang, Qian Cong, Lisa Kinch, Richard Schaeffer,
  Claudia Millán, Hahnbeom Park, Carson Adams, Caleb Glassman, Andy
  Degiovanni, Jose Pereira, Andria Rodrigues, Alberdina Dijk, Ana Ebrecht, and
  David Baker.
\newblock Accurate prediction of protein structures and interactions using a
  3-track network, 06 2021.

\bibitem[Baek et~al.(2020)Baek, Kim, Kim, and Lim]{baek2020autoclint}
Woonhyuk Baek, Ildoo Kim, Sungwoong Kim, and Sungbin Lim.
\newblock Autoclint: The winning method in autocv challenge 2019.
\newblock \emph{arXiv}, 2020.

\bibitem[Bai and An(2018)]{DBLP:journals/ijon/BaiA18}
Shuang Bai and Shan An.
\newblock A survey on automatic image caption generation.
\newblock \emph{Neurocomputing}, 311:\penalty0 291--304, 2018.
\newblock \doi{10.1016/j.neucom.2018.05.080}.
\newblock URL \url{https://doi.org/10.1016/j.neucom.2018.05.080}.

\bibitem[Bar{\'{o}} et~al.(2015)Bar{\'{o}}, Gonz{\`{a}}lez, Fabian, Bautista,
  Oliu, Escalante, Guyon, and Escalera]{DBLP:conf/cvpr/BaroGFBOEGE15}
Xavier Bar{\'{o}}, Jordi Gonz{\`{a}}lez, Junior Fabian, Miguel~{\'{A}}ngel
  Bautista, Marc Oliu, Hugo~Jair Escalante, Isabelle Guyon, and Sergio
  Escalera.
\newblock Chalearn looking at people 2015 challenges: Action spotting and
  cultural event recognition.
\newblock In \emph{2015 {IEEE} Conference on Computer Vision and Pattern
  Recognition Workshops, {CVPR} Workshops 2015, Boston, MA, USA, June 7-12,
  2015}, pages 1--9. {IEEE} Computer Society, 2015.
\newblock \doi{10.1109/CVPRW.2015.7301329}.
\newblock URL \url{https://doi.org/10.1109/CVPRW.2015.7301329}.

\bibitem[Black and Eskenazi(2009)]{black-eskenazi-2009-spoken}
Alan Black and Maxine Eskenazi.
\newblock The spoken dialogue challenge.
\newblock In \emph{Proceedings of the {SIGDIAL} 2009 Conference}, pages
  337--340, London, UK, September 2009. Association for Computational
  Linguistics.
\newblock URL \url{https://aclanthology.org/W09-3950}.

\bibitem[Braschler(2001)]{10.1007/3-540-44645-1_9}
Martin Braschler.
\newblock Clef 2000 --- overview of results.
\newblock In Carol Peters, editor, \emph{Cross-Language Information Retrieval
  and Evaluation}, pages 89--101, Berlin, Heidelberg, 2001. Springer Berlin
  Heidelberg.
\newblock ISBN 978-3-540-44645-3.

\bibitem[Carlens(2023)]{carlens2023state}
Harald Carlens.
\newblock State of competitive machine learning in 2022.
\newblock \emph{ML Contests Research}, 2023.
\newblock https://mlcontests.com/state-of-competitive-data-science-2022.

\bibitem[Carri{\'o}n-Ojeda et~al.(2022)Carri{\'o}n-Ojeda, Chen, Baz, Escalera,
  Guan, Guyon, Ullah, Wang, and Zhu]{https://doi.org/10.48550/arxiv.2208.14686}
Dustin Carri{\'o}n-Ojeda, Hong Chen, Adrian~El Baz, Sergio Escalera, Chaoyu
  Guan, Isabelle Guyon, Ihsan Ullah, Xin Wang, and Wenwu Zhu.
\newblock Neurips'22 cross-domain metadl competition: Design and baseline
  results, 2022.
\newblock URL \url{https://arxiv.org/abs/2208.14686}.

\bibitem[Clap{\'{e}}s et~al.(2020)Clap{\'{e}}s, J{\'{u}}nior, Morral, and
  Escalera]{DBLP:conf/fgr/ClapesJME20}
Albert Clap{\'{e}}s, J{\'{u}}lio C. S.~Jacques J{\'{u}}nior, Carla Morral, and
  Sergio Escalera.
\newblock Chalearn {LAP} 2020 challenge on identity-preserved human detection:
  Dataset and results.
\newblock In \emph{15th {IEEE} International Conference on Automatic Face and
  Gesture Recognition, {FG} 2020, Buenos Aires, Argentina, November 16-20,
  2020}, pages 801--808. {IEEE}, 2020.
\newblock \doi{10.1109/FG47880.2020.00135}.
\newblock URL \url{https://doi.org/10.1109/FG47880.2020.00135}.

\bibitem[Clough et~al.(2010)Clough, M{\"{u}}ller, and
  Sanderson]{DBLP:books/daglib/p/CloughMS10}
Paul~D. Clough, Henning M{\"{u}}ller, and Mark Sanderson.
\newblock Seven years of image retrieval evaluation.
\newblock In Henning M{\"{u}}ller, Paul~D. Clough, Thomas Deselaers, and
  Barbara Caputo, editors, \emph{ImageCLEF, Experimental Evaluation in Visual
  Information Retrieval}, pages 3--18. Springer, 2010.
\newblock \doi{10.1007/978-3-642-15181-1\_1}.
\newblock URL \url{https://doi.org/10.1007/978-3-642-15181-1\_1}.

\bibitem[Dosovitskiy et~al.(2017)Dosovitskiy, Ros, Codevilla, Lopez, and
  Koltun]{Dosovitskiy17}
Alexey Dosovitskiy, German Ros, Felipe Codevilla, Antonio Lopez, and Vladlen
  Koltun.
\newblock {CARLA}: {An} open urban driving simulator.
\newblock In \emph{Proceedings of the 1st Annual Conference on Robot Learning},
  pages 1--16, 2017.

\bibitem[Eichenberger et~al.(2022)Eichenberger, Neun, Martin, Herruzo,
  Spanring, Lu, Choi, Konyakhin, Lukashina, Shpilman, Wiedemann, Raubal, Wang,
  Vu, Mohajerpoor, Cai, Kim, Hermes, Melnik, Velioglu, Vieth, Schilling,
  Bojesomo, Marzouqi, Liatsis, Santokhi, Hillier, Yang, Sarwar, Jordan, Hewage,
  Jonietz, Tang, Gruca, Kopp, Kreil, and Hochreiter]{pmlr-v176-eichenberger22a}
Christian Eichenberger, Moritz Neun, Henry Martin, Pedro Herruzo, Markus
  Spanring, Yichao Lu, Sungbin Choi, Vsevolod Konyakhin, Nina Lukashina,
  Aleksei Shpilman, Nina Wiedemann, Martin Raubal, Bo~Wang, Hai~L. Vu, Reza
  Mohajerpoor, Chen Cai, Inhi Kim, Luca Hermes, Andrew Melnik, Riza Velioglu,
  Markus Vieth, Malte Schilling, Alabi Bojesomo, Hasan~Al Marzouqi, Panos
  Liatsis, Jay Santokhi, Dylan Hillier, Yiming Yang, Joned Sarwar, Anna Jordan,
  Emil Hewage, David Jonietz, Fei Tang, Aleksandra Gruca, Michael Kopp, David
  Kreil, and Sepp Hochreiter.
\newblock Traffic4cast at neurips 2021 - temporal and spatial few-shot transfer
  learning in gridded geo-spatial processes.
\newblock In Douwe Kiela, Marco Ciccone, and Barbara Caputo, editors,
  \emph{Proceedings of the NeurIPS 2021 Competitions and Demonstrations Track},
  volume 176 of \emph{Proceedings of Machine Learning Research}, pages 97--112.
  PMLR, 06--14 Dec 2022.
\newblock URL \url{https://proceedings.mlr.press/v176/eichenberger22a.html}.

\bibitem[El~Baz et~al.(2021{\natexlab{a}})El~Baz, Guyon, Liu, van Rijn,
  Treguer, and Vanschoren]{elbaz2021metadl}
Adrian El~Baz, Isabelle Guyon, Zhengying Liu, Jan~N. van Rijn, Sebastien
  Treguer, and Joaquin Vanschoren.
\newblock Metadl challenge design and baseline results.
\newblock In \emph{AAAI Workshop on Meta-Learning and MetaDL Challenge}, volume
  140 of \emph{Proceedings of Machine Learning Research}, pages 1--16. PMLR,
  2021{\natexlab{a}}.

\bibitem[El~Baz et~al.(2021{\natexlab{b}})El~Baz, Ullah, Alcoba\c{c}a,
  Carvalho, Chen, Ferreira, Gouk, Guan, Guyon, Hospedales, Hu, Huisman, Hutter,
  Liu, Mohr, {\"O}zt{\"u}rk, van Rijn, Sun, Wang, and Zhu]{elbaz:hal-03688638}
Adrian El~Baz, Ihsan Ullah, Edesio Alcoba\c{c}a, Andr{\'e} C. P. L.~F.
  Carvalho, Hong Chen, Fabio Ferreira, Henry Gouk, Chaoyu Guan, Isabelle Guyon,
  Timothy Hospedales, Shell Hu, Mike Huisman, Frank Hutter, Zhengying Liu,
  Felix Mohr, Ekrem {\"O}zt{\"u}rk, Jan~N van Rijn, Haozhe Sun, Xin Wang, and
  Wenwu Zhu.
\newblock {Lessons learned from the NeurIPS 2021 MetaDL challenge: Backbone
  fine-tuning without episodic meta-learning dominates for few-shot learning
  image classification}.
\newblock In \emph{{NeurIPS 2021 Competition and Demonstration Track}},
  On-line, United States, December 2021{\natexlab{b}}.
\newblock URL \url{https://hal.archives-ouvertes.fr/hal-03688638}.

\bibitem[Escalante and Hadsell(2019)]{DBLP:conf/nips/EscalanteH19}
Hugo~Jair Escalante and Raia Hadsell.
\newblock Neurips 2019 competition and demonstration track revised selected
  papers.
\newblock In Hugo~Jair Escalante and Raia Hadsell, editors, \emph{NeurIPS 2019
  Competition and Demonstration Track, 8-14 December 2019, Vancouver, Canada.
  Revised selected papers}, volume 123 of \emph{Proceedings of Machine Learning
  Research}, pages 1--12. {PMLR}, 2019.
\newblock URL \url{http://proceedings.mlr.press/v123/escalante20a.html}.

\bibitem[Escalante and Hofmann(2020)]{DBLP:conf/nips/EscalanteH20}
Hugo~Jair Escalante and Katja Hofmann.
\newblock Neurips 2020 competition and demonstration track: Revised selected
  papers.
\newblock In Hugo~Jair Escalante and Katja Hofmann, editors, \emph{NeurIPS 2020
  Competition and Demonstration Track, 6-12 December 2020, Virtual Event /
  Vancouver, BC, Canada}, volume 133 of \emph{Proceedings of Machine Learning
  Research}, pages 1--2. {PMLR}, 2020.
\newblock URL \url{http://proceedings.mlr.press/v133/escalante21a.html}.

\bibitem[Escalante et~al.(2010)Escalante, Hern{\'{a}}ndez, Gonz{\'{a}}lez,
  L{\'{o}}pez{-}L{\'{o}}pez, Montes{-}y{-}G{\'{o}}mez, Morales, Sucar, Pineda,
  and Grubinger]{DBLP:journals/cviu/EscalanteHGLMMSPG10}
Hugo~Jair Escalante, Carlos~A. Hern{\'{a}}ndez, Jes{\'{u}}s~A. Gonz{\'{a}}lez,
  Aurelio L{\'{o}}pez{-}L{\'{o}}pez, Manuel Montes{-}y{-}G{\'{o}}mez,
  Eduardo~F. Morales, Luis~Enrique Sucar, Luis~Villase{\~{n}}or Pineda, and
  Michael Grubinger.
\newblock The segmented and annotated {IAPR} {TC-12} benchmark.
\newblock \emph{Comput. Vis. Image Underst.}, 114\penalty0 (4):\penalty0
  419--428, 2010.
\newblock \doi{10.1016/j.cviu.2009.03.008}.
\newblock URL \url{https://doi.org/10.1016/j.cviu.2009.03.008}.

\bibitem[Escalante et~al.(2017)Escalante, Guyon, Escalera, J{\'{u}}nior,
  Madadi, Bar{\'{o}}, Ayache, Viegas, G{\"{u}}{\c{c}}l{\"{u}}t{\"{u}}rk,
  G{\"{u}}{\c{c}}l{\"{u}}, van Gerven, and van
  Lier]{DBLP:conf/ijcnn/EscalanteGEJMBA17}
Hugo~Jair Escalante, Isabelle Guyon, Sergio Escalera, J{\'{u}}lio C. S.~Jacques
  J{\'{u}}nior, Meysam Madadi, Xavier Bar{\'{o}}, St{\'{e}}phane Ayache,
  Evelyne Viegas, Yagmur G{\"{u}}{\c{c}}l{\"{u}}t{\"{u}}rk, Umut
  G{\"{u}}{\c{c}}l{\"{u}}, Marcel A.~J. van Gerven, and Rob van Lier.
\newblock Design of an explainable machine learning challenge for video
  interviews.
\newblock In \emph{2017 International Joint Conference on Neural Networks,
  {IJCNN} 2017, Anchorage, AK, USA, May 14-19, 2017}, pages 3688--3695. {IEEE},
  2017.
\newblock \doi{10.1109/IJCNN.2017.7966320}.
\newblock URL \url{https://doi.org/10.1109/IJCNN.2017.7966320}.

\bibitem[Escalante et~al.(2019)Escalante, Tu, Guyon, Silver, Viegas, Chen, Dai,
  and Yang]{DBLP:journals/corr/abs-1903-05263}
Hugo~Jair Escalante, Wei{-}Wei Tu, Isabelle Guyon, Daniel~L. Silver, Evelyne
  Viegas, Yuqiang Chen, Wenyuan Dai, and Qiang Yang.
\newblock Automl @ neurips 2018 challenge: Design and results.
\newblock \emph{CoRR}, abs/1903.05263, 2019.
\newblock URL \url{http://arxiv.org/abs/1903.05263}.

\bibitem[Escalante et~al.(2022)Escalante, Kaya, Salah, Escalera,
  G{\"{u}}{\c{c}}l{\"{u}}t{\"{u}}rk, G{\"{u}}{\c{c}}l{\"{u}}, Bar{\'{o}},
  Guyon, J{\'{u}}nior, Madadi, Ayache, Viegas, G{\"{u}}rpinar, Wicaksana, Liem,
  van Gerven, and van Lier]{DBLP:journals/taffco/EscalanteKSEGGB22}
Hugo~Jair Escalante, Heysem Kaya, Albert~Ali Salah, Sergio Escalera, Yagmur
  G{\"{u}}{\c{c}}l{\"{u}}t{\"{u}}rk, Umut G{\"{u}}{\c{c}}l{\"{u}}, Xavier
  Bar{\'{o}}, Isabelle Guyon, J{\'{u}}lio C. S.~Jacques J{\'{u}}nior, Meysam
  Madadi, St{\'{e}}phane Ayache, Evelyne Viegas, Furkan G{\"{u}}rpinar,
  Achmadnoer~Sukma Wicaksana, Cynthia C.~S. Liem, Marcel A.~J. van Gerven, and
  Rob van Lier.
\newblock Modeling, recognizing, and explaining apparent personality from
  videos.
\newblock \emph{{IEEE} Trans. Affect. Comput.}, 13\penalty0 (2):\penalty0
  894--911, 2022.
\newblock \doi{10.1109/TAFFC.2020.2973984}.
\newblock URL \url{https://doi.org/10.1109/TAFFC.2020.2973984}.

\bibitem[Escalera et~al.(2013)Escalera, Gonz{\`{a}}lez, Bar{\'{o}}, Reyes,
  Lopes, Guyon, Athitsos, and Escalante]{DBLP:conf/icmi/EscaleraGBRLGAE13}
Sergio Escalera, Jordi Gonz{\`{a}}lez, Xavier Bar{\'{o}}, Miguel Reyes, Oscar
  Lopes, Isabelle Guyon, Vassilis Athitsos, and Hugo~Jair Escalante.
\newblock Multi-modal gesture recognition challenge 2013: dataset and results.
\newblock In Julien Epps, Fang Chen, Sharon~L. Oviatt, Kenji Mase, Andrew
  Sears, Kristiina Jokinen, and Bj{\"{o}}rn~W. Schuller, editors, \emph{2013
  International Conference on Multimodal Interaction, {ICMI} '13, Sydney, NSW,
  Australia, December 9-13, 2013}, pages 445--452. {ACM}, 2013.
\newblock \doi{10.1145/2522848.2532595}.
\newblock URL \url{https://doi.org/10.1145/2522848.2532595}.

\bibitem[Escalera et~al.(2014)Escalera, Bar{\'{o}}, Gonz{\`{a}}lez, Bautista,
  Madadi, Reyes, Ponce{-}L{\'{o}}pez, Escalante, Shotton, and
  Guyon]{DBLP:conf/eccv/EscaleraBGBMRPE14}
Sergio Escalera, Xavier Bar{\'{o}}, Jordi Gonz{\`{a}}lez, Miguel~{\'{A}}ngel
  Bautista, Meysam Madadi, Miguel Reyes, V{\'{\i}}ctor Ponce{-}L{\'{o}}pez,
  Hugo~Jair Escalante, Jamie Shotton, and Isabelle Guyon.
\newblock Chalearn looking at people challenge 2014: Dataset and results.
\newblock In Lourdes Agapito, Michael~M. Bronstein, and Carsten Rother,
  editors, \emph{Computer Vision - {ECCV} 2014 Workshops - Zurich, Switzerland,
  September 6-7 and 12, 2014, Proceedings, Part {I}}, volume 8925 of
  \emph{Lecture Notes in Computer Science}, pages 459--473. Springer, 2014.
\newblock \doi{10.1007/978-3-319-16178-5\_32}.
\newblock URL \url{https://doi.org/10.1007/978-3-319-16178-5\_32}.

\bibitem[Escalera et~al.(2015)Escalera, Fabian, Pardo, Bar{\'{o}},
  Gonz{\`{a}}lez, Escalante, Misevic, Steiner, and
  Guyon]{DBLP:conf/iccvw/EscaleraFPBGEMS15}
Sergio Escalera, Junior Fabian, Pablo Pardo, Xavier Bar{\'{o}}, Jordi
  Gonz{\`{a}}lez, Hugo~Jair Escalante, Dusan Misevic, Ulrich Steiner, and
  Isabelle Guyon.
\newblock Chalearn looking at people 2015: Apparent age and cultural event
  recognition datasets and results.
\newblock In \emph{2015 {IEEE} International Conference on Computer Vision
  Workshop, {ICCV} Workshops 2015, Santiago, Chile, December 7-13, 2015}, pages
  243--251. {IEEE} Computer Society, 2015.
\newblock \doi{10.1109/ICCVW.2015.40}.
\newblock URL \url{https://doi.org/10.1109/ICCVW.2015.40}.

\bibitem[Escalera et~al.(2016)Escalera, Torres, Mart{\'{\i}}nez, Bar{\'{o}},
  Escalante, Guyon, Tzimiropoulos, Corneanu, Oliu, Bagheri, and
  Valstar]{DBLP:conf/cvpr/EscaleraTMBEGTC16}
Sergio Escalera, Mercedes Torres, Brais Mart{\'{\i}}nez, Xavier Bar{\'{o}},
  Hugo~Jair Escalante, Isabelle Guyon, Georgios Tzimiropoulos, Ciprian~A.
  Corneanu, Marc Oliu, Mohammad~Ali Bagheri, and Michel~F. Valstar.
\newblock Chalearn looking at people and faces of the world: Face
  analysisworkshop and challenge 2016.
\newblock In \emph{2016 {IEEE} Conference on Computer Vision and Pattern
  Recognition Workshops, {CVPR} Workshops 2016, Las Vegas, NV, USA, June 26 -
  July 1, 2016}, pages 706--713. {IEEE} Computer Society, 2016.
\newblock \doi{10.1109/CVPRW.2016.93}.
\newblock URL \url{https://doi.org/10.1109/CVPRW.2016.93}.

\bibitem[Escalera et~al.(2017{\natexlab{a}})Escalera, Bar{\'{o}}, Escalante,
  and Guyon]{DBLP:conf/ijcnn/EscaleraBEG17}
Sergio Escalera, Xavier Bar{\'{o}}, Hugo~Jair Escalante, and Isabelle Guyon.
\newblock Chalearn looking at people: {A} review of events and resources.
\newblock In \emph{2017 International Joint Conference on Neural Networks,
  {IJCNN} 2017, Anchorage, AK, USA, May 14-19, 2017}, pages 1594--1601. {IEEE},
  2017{\natexlab{a}}.
\newblock \doi{10.1109/IJCNN.2017.7966041}.
\newblock URL \url{https://doi.org/10.1109/IJCNN.2017.7966041}.

\bibitem[Escalera et~al.(2017{\natexlab{b}})Escalera, Guyon, and
  Athitsos]{DBLP:books/sp/EGA2017}
Sergio Escalera, Isabelle Guyon, and Vassilis Athitsos, editors.
\newblock \emph{Gesture Recognition}.
\newblock Springer, 2017{\natexlab{b}}.
\newblock ISBN 978-3-319-57020-4.
\newblock \doi{10.1007/978-3-319-57021-1}.
\newblock URL \url{https://doi.org/10.1007/978-3-319-57021-1}.

\bibitem[Evans et~al.(2018)Evans, Jumper, Kirkpatrick, Sifre, Green, Qin,
  Žídek, Nelson, Bridgland, Penedones, Petersen, Simonyan, Crossan, Jones,
  Silver, Kavukcuoglu, Hassabis, and Senior]{casp13}
Richard Evans, John Jumper, James Kirkpatrick, Laurent Sifre, Tim Green,
  Chongli Qin, Augustin Žídek, Sandy Nelson, Alex Bridgland, Hugo Penedones,
  Stig Petersen, Karen Simonyan, Steve Crossan, David Jones, David Silver,
  Koray Kavukcuoglu, Demis Hassabis, and Andrew Senior.
\newblock De novo structure prediction with deep-learning based scoring, 12
  2018.

\bibitem[Everingham et~al.(2015)Everingham, Eslami, Van~Gool, Williams, Winn,
  and Zisserman]{Everingham15}
M.~Everingham, S.~M.~A. Eslami, L.~Van~Gool, C.~K.~I. Williams, J.~Winn, and
  A.~Zisserman.
\newblock The pascal visual object classes challenge: A retrospective.
\newblock \emph{International Journal of Computer Vision}, 111\penalty0
  (1):\penalty0 98--136, January 2015.

\bibitem[Feurer et~al.(2019)Feurer, Klein, Eggensperger, Springenberg, Blum,
  and Hutter]{Feurer2019}
Matthias Feurer, Aaron Klein, Katharina Eggensperger, Jost~Tobias Springenberg,
  Manuel Blum, and Frank Hutter.
\newblock \emph{Auto-sklearn: Efficient and Robust Automated Machine Learning},
  pages 113--134.
\newblock Springer International Publishing, Cham, 2019.
\newblock ISBN 978-3-030-05318-5.
\newblock \doi{10.1007/978-3-030-05318-5_6}.
\newblock URL \url{https://doi.org/10.1007/978-3-030-05318-5_6}.

\bibitem[Gardner et~al.(2020)Gardner, Lowman, Richardson, Llorens, Markowitz,
  Drenkow, Newman, Clark, Perrotta, Perrotta, Highley, Shcherbina, Bernadoni,
  Jordan, and Asenov]{pmlr-v123-gardner20a}
Ryan~W. Gardner, Corey Lowman, Casey Richardson, Ashley~J. Llorens, Jared
  Markowitz, Nathan Drenkow, Andrew Newman, Gregory Clark, Gino Perrotta,
  Robert Perrotta, Timothy Highley, Vlad Shcherbina, William Bernadoni, Mark
  Jordan, and Asen Asenov.
\newblock The first international competition in machine reconnaissance blind
  chess.
\newblock In Hugo~Jair Escalante and Raia Hadsell, editors, \emph{Proceedings
  of the NeurIPS 2019 Competition and Demonstration Track}, volume 123 of
  \emph{Proceedings of Machine Learning Research}, pages 121--130. PMLR, 08--14
  Dec 2020.
\newblock URL \url{https://proceedings.mlr.press/v123/gardner20a.html}.

\bibitem[Garofolo et~al.(2000)Garofolo, Auzanne, and
  Voorhees]{10.5555/2835865.2835867}
John~S. Garofolo, Cedric G.~P. Auzanne, and Ellen~M. Voorhees.
\newblock The trec spoken document retrieval track: A success story.
\newblock In \emph{Content-Based Multimedia Information Access - Volume 1},
  RIAO '00, page 1–20, Paris, FRA, 2000. LE CENTRE DE HAUTES ETUDES
  INTERNATIONALES D'INFORMATIQUE DOCUMENTAIRE.

\bibitem[Garris et~al.(1997)Garris, Blue, Candela, Grother, Janet, and
  Wilson]{mnistreport97}
Michael Garris, J~Blue, Gerald Candela, Patrick Grother, Stanley Janet, and
  Charles Wilson.
\newblock Nist form-based handprint recognition system, 1997-01-01 1997.

\bibitem[Grother(1995)]{Grother1995NISTSD}
Patrick Grother.
\newblock Nist special database 19 handprinted forms and characters database,
  1995.

\bibitem[Gu et~al.(2022)Gu, Cai, Dong, Ren, and Timofte]{ntire22}
Jinjin Gu, Haoming Cai, Chao Dong, Jimmy~S. Ren, and Radu Timofte.
\newblock Ntire 2022 challenge on perceptual image quality assessment, 2022.
\newblock URL \url{https://arxiv.org/abs/2206.11695}.

\bibitem[Guss et~al.(2021)Guss, Milani, Topin, Houghton, Mohanty, Melnik,
  Harter, Buschmaas, Jaster, Berganski, Heitkamp, Henning, Ritter, Wu, Hao, Lu,
  Mao, Mao, Wang, Opanowicz, Kanervisto, Schraner, Scheller, Zhou, Liu, Nishio,
  Tsuneda, Ramanauskas, and Juceviciute]{pmlr-v133-guss21a}
William~Hebgen Guss, Stephanie Milani, Nicholay Topin, Brandon Houghton,
  Sharada Mohanty, Andrew Melnik, Augustin Harter, Benoit Buschmaas, Bjarne
  Jaster, Christoph Berganski, Dennis Heitkamp, Marko Henning, Helge Ritter,
  Chengjie Wu, Xiaotian Hao, Yiming Lu, Hangyu Mao, Yihuan Mao, Chao Wang,
  Michal Opanowicz, Anssi Kanervisto, Yanick Schraner, Christian Scheller,
  Xiren Zhou, Lu~Liu, Daichi Nishio, Toi Tsuneda, Karolis Ramanauskas, and
  Gabija Juceviciute.
\newblock Towards robust and domain agnostic reinforcement learning
  competitions: Minerl 2020.
\newblock In Hugo~Jair Escalante and Katja Hofmann, editors, \emph{Proceedings
  of the NeurIPS 2020 Competition and Demonstration Track}, volume 133 of
  \emph{Proceedings of Machine Learning Research}, pages 233--252. PMLR, 06--12
  Dec 2021.
\newblock URL \url{https://proceedings.mlr.press/v133/guss21a.html}.

\bibitem[Guyon et~al.(2004)Guyon, Gunn, Ben-Hur, and Dror]{NIPS2004_5e751896}
Isabelle Guyon, Steve Gunn, Asa Ben-Hur, and Gideon Dror.
\newblock Result analysis of the nips 2003 feature selection challenge.
\newblock In L.~Saul, Y.~Weiss, and L.~Bottou, editors, \emph{Advances in
  Neural Information Processing Systems}, volume~17. MIT Press, 2004.
\newblock URL
  \url{https://proceedings.neurips.cc/paper/2004/file/5e751896e527c862bf67251a474b3819-Paper.pdf}.

\bibitem[Guyon et~al.(2006)Guyon, Alamdari, Dror, and
  Buhmann]{DBLP:conf/ijcnn/GuyonADB06}
Isabelle Guyon, Amir Reza Saffari~Azar Alamdari, Gideon Dror, and Joachim~M.
  Buhmann.
\newblock Performanceprediction challenge.
\newblock In \emph{Proceedings of the International Joint Conference on Neural
  Networks, {IJCNN} 2006, part of the {IEEE} World Congress on Computational
  Intelligence, {WCCI} 2006, Vancouver, BC, Canada, 16-21 July 2006}, pages
  1649--1656. {IEEE}, 2006.
\newblock \doi{10.1109/IJCNN.2006.246632}.
\newblock URL \url{https://doi.org/10.1109/IJCNN.2006.246632}.

\bibitem[Guyon et~al.(2008)Guyon, Saffari, Dror, and
  Cawley]{DBLP:journals/nn/GuyonSDC08}
Isabelle Guyon, Amir Saffari, Gideon Dror, and Gavin~C. Cawley.
\newblock Analysis of the {IJCNN} 2007 agnostic learning vs. prior knowledge
  challenge.
\newblock \emph{Neural Networks}, 21\penalty0 (2-3):\penalty0 544--550, 2008.
\newblock \doi{10.1016/j.neunet.2007.12.024}.
\newblock URL \url{https://doi.org/10.1016/j.neunet.2007.12.024}.

\bibitem[Guyon et~al.(2019)Guyon, Sun{-}Hosoya, Boull{\'{e}}, Escalante,
  Escalera, Liu, Jajetic, Ray, Saeed, Sebag, Statnikov, Tu, and
  Viegas]{DBLP:books/sp/19/GuyonSBEELJRSSSTV19}
Isabelle Guyon, Lisheng Sun{-}Hosoya, Marc Boull{\'{e}}, Hugo~Jair Escalante,
  Sergio Escalera, Zhengying Liu, Damir Jajetic, Bisakha Ray, Mehreen Saeed,
  Mich{\'{e}}le Sebag, Alexander~R. Statnikov, Wei{-}Wei Tu, and Evelyne
  Viegas.
\newblock Analysis of the automl challenge series 2015-2018.
\newblock In Frank Hutter, Lars Kotthoff, and Joaquin Vanschoren, editors,
  \emph{Automated Machine Learning - Methods, Systems, Challenges}, The
  Springer Series on Challenges in Machine Learning, pages 177--219. Springer,
  2019.
\newblock \doi{10.1007/978-3-030-05318-5\_10}.
\newblock URL \url{https://doi.org/10.1007/978-3-030-05318-5\_10}.

\bibitem[Harman(1993)]{trec93}
Donna Harman.
\newblock Overview of the first trec conference.
\newblock In \emph{Proceedings of the 16th Annual International ACM SIGIR
  Conference on Research and Development in Information Retrieval}, SIGIR '93,
  page 36–47, New York, NY, USA, 1993. Association for Computing Machinery.
\newblock ISBN 0897916050.
\newblock \doi{10.1145/160688.160692}.
\newblock URL \url{https://doi.org/10.1145/160688.160692}.

\bibitem[Harman(1995)]{DBLP:conf/trec/Harman95}
Donna Harman.
\newblock Overview of the fourth text retrieval conference {(TREC-4)}.
\newblock In Donna~K. Harman, editor, \emph{Proceedings of The Fourth Text
  REtrieval Conference, {TREC} 1995, Gaithersburg, Maryland, USA, November 1-3,
  1995}, volume 500-236 of \emph{{NIST} Special Publication}. National
  Institute of Standards and Technology {(NIST)}, 1995.
\newblock URL \url{http://trec.nist.gov/pubs/trec4/overview.ps.gz}.

\bibitem[Harman(1998)]{harman:trec}
Donna Harman.
\newblock {The Text REtrieval Conferences (TRECs) and the Cross-Language
  Track}.
\newblock In \emph{First International Conference on Language Resources \&
  Evaluation, Granada, Spain}, pages 517--522, 1998.

\bibitem[He et~al.(2015)He, Zhang, Ren, and Sun]{DBLP:journals/corr/HeZRS15}
Kaiming He, Xiangyu Zhang, Shaoqing Ren, and Jian Sun.
\newblock Deep residual learning for image recognition.
\newblock \emph{CoRR}, abs/1512.03385, 2015.
\newblock URL \url{http://arxiv.org/abs/1512.03385}.

\bibitem[Hofseth(2017)]{10.1093/carcin/bgx085}
Lorne~J Hofseth.
\newblock {Getting rigorous with scientific rigor}.
\newblock \emph{Carcinogenesis}, 39\penalty0 (1):\penalty0 21--25, 08 2017.
\newblock ISSN 0143-3334.
\newblock \doi{10.1093/carcin/bgx085}.
\newblock URL \url{https://doi.org/10.1093/carcin/bgx085}.

\bibitem[Hutter et~al.(2018)Hutter, Kotthoff, and Vanschoren]{automlbook}
Frank Hutter, Lars Kotthoff, and Joaquin Vanschoren.
\newblock \emph{AutoML: Methods, Systems, Challenges}.
\newblock Springer Series in Challenges in Machine Learning. Springer, 2018.

\bibitem[J et~al.(2014)J, K, A, T, and Tramontano]{CASP04}
Moult J, Fidelis K, Kryshtafovych A, Schwede T, and Tramontano.
\newblock Critical assessment of methods of protein structure prediction (casp)
  round {X}.
\newblock \emph{Proteins}, 2\penalty0 (2):\penalty0 1--6, 2014.

\bibitem[Ju et~al.(2021)Ju, Yang, Jia, Ye, Chen, Tan, Sun, Shi, and
  Ding]{DANet}
Bo~Ju, Wei Yang, Jinrang Jia, Xiaoqing Ye, Qu~Chen, Xiao Tan, Hao Sun, Yifeng
  Shi, and Errui Ding.
\newblock Danet: Dimension apart network for radar object detection.
\newblock In \emph{Proceedings of the 2021 International Conference on
  Multimedia Retrieval}, ICMR '21, page 533–539, New York, NY, USA, 2021.
  Association for Computing Machinery.
\newblock ISBN 9781450384636.
\newblock \doi{10.1145/3460426.3463656}.
\newblock URL \url{https://doi.org/10.1145/3460426.3463656}.

\bibitem[Jumper et~al.(2021)Jumper, Evans, Pritzel, Green, Figurnov,
  Ronneberger, Tunyasuvunakool, Bates, Z{\'i}dek, Potapenko, Bridgland, Meyer,
  Kohl, Ballard, Cowie, Romera-Paredes, Nikolov, Jain, Adler, Back, Petersen,
  Reiman, Clancy, Zielinski, Steinegger, Pacholska, Berghammer, Bodenstein,
  Silver, Vinyals, Senior, Kavukcuoglu, Kohli, and
  Hassabis]{Jumper2021HighlyAP}
John~M. Jumper, Richard Evans, Alexander Pritzel, Tim Green, Michael Figurnov,
  Olaf Ronneberger, Kathryn Tunyasuvunakool, Russ Bates, Augustin Z{\'i}dek,
  Anna Potapenko, Alex Bridgland, Clemens Meyer, Simon A~A Kohl, Andy Ballard,
  Andrew Cowie, Bernardino Romera-Paredes, Stanislav Nikolov, Rishub Jain,
  Jonas Adler, Trevor Back, Stig Petersen, David~A. Reiman, Ellen Clancy,
  Michal Zielinski, Martin Steinegger, Michalina Pacholska, Tamas Berghammer,
  Sebastian Bodenstein, David Silver, Oriol Vinyals, Andrew~W. Senior, Koray
  Kavukcuoglu, Pushmeet Kohli, and Demis Hassabis.
\newblock Highly accurate protein structure prediction with alphafold.
\newblock \emph{Nature}, 596:\penalty0 583 -- 589, 2021.

\bibitem[Kaufman et~al.(2011)Kaufman, Rosset, and Perlich]{LeakageSIGKDD}
Shachar Kaufman, Saharon Rosset, and Claudia Perlich.
\newblock Leakage in data mining: Formulation, detection, and avoidance.
\newblock In \emph{Proceedings of the ACM SIGKDD International Conference on
  Knowledge Discovery and Data Mining}, volume~6, pages 556--563, 01 2011.
\newblock \doi{10.1145/2020408.2020496}.

\bibitem[Kidzinski et~al.(2018)Kidzinski, Mohanty, Ong, Huang, Zhou, Pechenko,
  Stelmaszczyk, Jarosik, Pavlov, Kolesnikov, Plis, Chen, Zhang, Chen, Shi,
  Zheng, Yuan, Lin, Michalewski, Milos, Osinski, Melnik, Schilling, Ritter,
  Carroll, Hicks, Levine, Salath{\'{e}}, and
  Delp]{DBLP:journals/corr/abs-1804-00361}
Lukasz Kidzinski, Sharada~Prasanna Mohanty, Carmichael~F. Ong, Zhewei Huang,
  Shuchang Zhou, Anton Pechenko, Adam Stelmaszczyk, Piotr Jarosik, Mikhail
  Pavlov, Sergey Kolesnikov, Sergey~M. Plis, Zhibo Chen, Zhizheng Zhang, Jiale
  Chen, Jun Shi, Zhuobin Zheng, Chun Yuan, Zhihui Lin, Henryk Michalewski,
  Piotr Milos, Blazej Osinski, Andrew Melnik, Malte Schilling, Helge~J. Ritter,
  Sean~F. Carroll, Jennifer~L. Hicks, Sergey Levine, Marcel Salath{\'{e}}, and
  Scott~L. Delp.
\newblock Learning to run challenge solutions: Adapting reinforcement learning
  methods for neuromusculoskeletal environments.
\newblock \emph{CoRR}, abs/1804.00361, 2018.
\newblock URL \url{http://arxiv.org/abs/1804.00361}.

\bibitem[Kiela et~al.(2022)Kiela, Ciccone, and Caputo]{pmlr-v176-kiela22a}
Douwe Kiela, Marco Ciccone, and Barbara Caputo.
\newblock Neurips 2021 competition and demonstration track revised selected
  papers.
\newblock In Douwe Kiela, Marco Ciccone, and Barbara Caputo, editors,
  \emph{Proceedings of the NeurIPS 2021 Competitions and Demonstrations Track},
  volume 176 of \emph{Proceedings of Machine Learning Research}, pages i--ii.
  PMLR, 06--14 Dec 2022.
\newblock URL \url{https://proceedings.mlr.press/v176/kiela22a.html}.

\bibitem[Kitano et~al.(1998)Kitano, Tambe, Stone, Veloso, Coradeschi, Osawa,
  Matsubara, Noda, and Asada]{robocup1}
Hiroaki Kitano, Milind Tambe, Peter Stone, Manuela Veloso, Silvia Coradeschi,
  Eiichi Osawa, Hitoshi Matsubara, Itsuki Noda, and Minoru Asada.
\newblock The robocup synthetic agent challenge 97.
\newblock In Hiroaki Kitano, editor, \emph{RoboCup-97: Robot Soccer World Cup
  I}, pages 62--73, Berlin, Heidelberg, 1998. Springer Berlin Heidelberg.
\newblock ISBN 978-3-540-69789-3.

\bibitem[Kopp et~al.(2021)Kopp, Kreil, Neun, Jonietz, Martin, Herruzo, Gruca,
  Soleymani, Wu, Liu, Xu, Zhang, Santokhi, Bojesomo, Marzouqi, Liatsis, Kwok,
  Qi, and Hochreiter]{pmlr-v133-kopp21a}
Michael Kopp, David Kreil, Moritz Neun, David Jonietz, Henry Martin, Pedro
  Herruzo, Aleksandra Gruca, Ali Soleymani, Fanyou Wu, Yang Liu, Jingwei Xu,
  Jianjin Zhang, Jay Santokhi, Alabi Bojesomo, Hasan~Al Marzouqi, Panos
  Liatsis, Pak~Hay Kwok, Qi~Qi, and Sepp Hochreiter.
\newblock Traffic4cast at neurips 2020 - yet more on the unreasonable
  effectiveness of gridded geo-spatial processes.
\newblock In Hugo~Jair Escalante and Katja Hofmann, editors, \emph{Proceedings
  of the NeurIPS 2020 Competition and Demonstration Track}, volume 133 of
  \emph{Proceedings of Machine Learning Research}, pages 325--343. PMLR, 06--12
  Dec 2021.
\newblock URL \url{https://proceedings.mlr.press/v133/kopp21a.html}.

\bibitem[Koren(2009)]{bellkor}
Y.~Koren.
\newblock The bellkor solution to the netflix grand prize.
\newblock Netflix prize documentation 81, 1-10, 2009.

\bibitem[Kreil et~al.(2020)Kreil, Kopp, Jonietz, Neun, Gruca, Herruzo, Martin,
  Soleymani, and Hochreiter]{pmlr-v123-kreil20a}
David~P Kreil, Michael~K Kopp, David Jonietz, Moritz Neun, Aleksandra Gruca,
  Pedro Herruzo, Henry Martin, Ali Soleymani, and Sepp Hochreiter.
\newblock The surprising efficiency of framing geo-spatial time series
  forecasting as a video prediction task – insights from the iarai \t4c
  competition at neurips 2019.
\newblock In Hugo~Jair Escalante and Raia Hadsell, editors, \emph{Proceedings
  of the NeurIPS 2019 Competition and Demonstration Track}, volume 123 of
  \emph{Proceedings of Machine Learning Research}, pages 232--241. PMLR, 08--14
  Dec 2020.
\newblock URL \url{https://proceedings.mlr.press/v123/kreil20a.html}.

\bibitem[Krizhevsky et~al.(2012)Krizhevsky, Sutskever, and
  Hinton]{DBLP:conf/nips/KrizhevskySH12}
Alex Krizhevsky, Ilya Sutskever, and Geoffrey~E. Hinton.
\newblock Imagenet classification with deep convolutional neural networks.
\newblock In Peter~L. Bartlett, Fernando C.~N. Pereira, Christopher J.~C.
  Burges, L{\'{e}}on Bottou, and Kilian~Q. Weinberger, editors, \emph{Advances
  in Neural Information Processing Systems 25: 26th Annual Conference on Neural
  Information Processing Systems 2012. Proceedings of a meeting held December
  3-6, 2012, Lake Tahoe, Nevada, United States}, pages 1106--1114, 2012.
\newblock URL
  \url{https://proceedings.neurips.cc/paper/2012/hash/c399862d3b9d6b76c8436e924a68c45b-Abstract.html}.

\bibitem[Kryshtafovych et~al.(2021)Kryshtafovych, Schwede, Topf, Fidelis, and
  Moult]{CASP21}
Andriy Kryshtafovych, Torsten Schwede, Maya Topf, Krzysztof Fidelis, and John
  Moult.
\newblock Critical assessment of methods of protein structure prediction
  (casp)—round {XIV}.
\newblock \emph{Proteins}, 89\penalty0 (12):\penalty0 1607--1617, 2021.

\bibitem[Li et~al.(2020)Li, Han, Lu, Niu, Yu, Dantcheva, Zhao, and
  Shan]{repss1}
Xiaobai Li, Hu~Han, Hao Lu, Xuesong Niu, Zitong Yu, Antitza Dantcheva, Guoying
  Zhao, and Shiguang Shan.
\newblock The 1st challenge on remote physiological signal sensing (repss),
  2020.
\newblock URL \url{https://arxiv.org/abs/2003.11756}.

\bibitem[Li et~al.(2021)Li, Sun, Sun, Han, Dantcheva, Shan, and Zhao]{repss2}
Xiaobai Li, Haomiao Sun, Zhaodong Sun, Hu~Han, Antitza Dantcheva, Shiguang
  Shan, and Guoying Zhao.
\newblock The 2nd challenge on remote physiological signal sensing (repss).
\newblock In \emph{2021 IEEE/CVF International Conference on Computer Vision
  Workshops (ICCVW)}, pages 2404--2413, 2021.
\newblock \doi{10.1109/ICCVW54120.2021.00273}.

\bibitem[Lin et~al.(2014)Lin, Maire, Belongie, Hays, Perona, Ramanan,
  Doll{\'{a}}r, and Zitnick]{DBLP:conf/eccv/LinMBHPRDZ14}
Tsung{-}Yi Lin, Michael Maire, Serge~J. Belongie, James Hays, Pietro Perona,
  Deva Ramanan, Piotr Doll{\'{a}}r, and C.~Lawrence Zitnick.
\newblock Microsoft {COCO:} common objects in context.
\newblock In David~J. Fleet, Tom{\'{a}}s Pajdla, Bernt Schiele, and Tinne
  Tuytelaars, editors, \emph{Computer Vision - {ECCV} 2014 - 13th European
  Conference, Zurich, Switzerland, September 6-12, 2014, Proceedings, Part
  {V}}, volume 8693 of \emph{Lecture Notes in Computer Science}, pages
  740--755. Springer, 2014.
\newblock \doi{10.1007/978-3-319-10602-1\_48}.
\newblock URL \url{https://doi.org/10.1007/978-3-319-10602-1\_48}.

\bibitem[Liu et~al.(2019)Liu, Wan, Escalera, Escalante, Tan, Yuan, Wang, Lin,
  Guo, Guyon, and Li]{DBLP:conf/cvpr/Liu0EETYWLGGL19}
Ajian Liu, Jun Wan, Sergio Escalera, Hugo~Jair Escalante, Zichang Tan, Qi~Yuan,
  Kai Wang, Chi Lin, Guodong Guo, Isabelle Guyon, and Stan~Z. Li.
\newblock Multi-modal face anti-spoofing attack detection challenge at
  {CVPR2019}.
\newblock In \emph{{IEEE} Conference on Computer Vision and Pattern Recognition
  Workshops, {CVPR} Workshops 2019, Long Beach, CA, USA, June 16-20, 2019},
  pages 1601--1610. Computer Vision Foundation / {IEEE}, 2019.
\newblock \doi{10.1109/CVPRW.2019.00202}.
\newblock URL
  \url{http://openaccess.thecvf.com/content\_CVPRW\_2019/html/CFS/Liu\_Multi-Modal\_Face\_Anti-Spoofing\_Attack\_Detection\_Challenge\_at\_CVPR2019\_CVPRW\_2019\_paper.html}.

\bibitem[Liu et~al.(2021{\natexlab{a}})Liu, Li, Wan, Liang, Escalera,
  Escalante, Madadi, Jin, Wu, Yu, Tan, Yuan, Yang, Zhou, Guo, and
  Li]{DBLP:journals/iet-bmt/LiuL0LEEM0WYTYY21}
Ajian Liu, Xuan Li, Jun Wan, Yanyan Liang, Sergio Escalera, Hugo~Jair
  Escalante, Meysam Madadi, Yi~Jin, Zhuoyuan Wu, Xiaogang Yu, Zichang Tan,
  Qi~Yuan, Ruikun Yang, Benjia Zhou, Guodong Guo, and Stan~Z. Li.
\newblock Cross-ethnicity face anti-spoofing recognition challenge: {A} review.
\newblock \emph{{IET} Biom.}, 10\penalty0 (1):\penalty0 24--43,
  2021{\natexlab{a}}.
\newblock \doi{10.1049/bme2.12002}.
\newblock URL \url{https://doi.org/10.1049/bme2.12002}.

\bibitem[Liu et~al.(2021{\natexlab{b}})Liu, Pavao, Xu, Escalera, Ferreira,
  Guyon, Hong, Hutter, Ji, Junior, Li, Lindauer, Luo, Madadi, Nierhoff, Niu,
  Pan, Stoll, Treguer, Wang, Wang, Wu, Xiong, Zela, and
  Zhang]{ChaLearnAutoDL2019}
Zhengying Liu, Adrien Pavao, Zhen Xu, Sergio Escalera, Fabio Ferreira, Isabelle
  Guyon, Sirui Hong, Frank Hutter, Rongrong Ji, Julio C. S.~Jacques Junior,
  Ge~Li, Marius Lindauer, Zhipeng Luo, Meysam Madadi, Thomas Nierhoff, Kangning
  Niu, Chunguang Pan, Danny Stoll, Sebastien Treguer, Jin Wang, Peng Wang,
  Chenglin Wu, Youcheng Xiong, Arbër Zela, and Yang Zhang.
\newblock Winning solutions and post-challenge analyses of the chalearn autodl
  challenge 2019.
\newblock \emph{IEEE Transactions on Pattern Analysis and Machine
  Intelligence}, 43\penalty0 (9):\penalty0 3108--3125, 2021{\natexlab{b}}.
\newblock \doi{10.1109/TPAMI.2021.3075372}.

\bibitem[Lucas et~al.(2003)Lucas, Panaretos, Sosa, Tang, Wong, and
  Young]{DBLP:conf/icdar/LucasPSTWY03}
Simon~M. Lucas, Alex Panaretos, Luis Sosa, Anthony Tang, Shirley Wong, and
  Robert Young.
\newblock {ICDAR} 2003 robust reading competitions.
\newblock In \emph{7th International Conference on Document Analysis and
  Recognition {(ICDAR} 2003), 2-Volume Set, 3-6 August 2003, Edinburgh,
  Scotland, {UK}}, pages 682--687. {IEEE} Computer Society, 2003.
\newblock \doi{10.1109/ICDAR.2003.1227749}.
\newblock URL \url{https://doi.org/10.1109/ICDAR.2003.1227749}.

\bibitem[Marak et~al.(2009)Marak, Cousty, Najman, and
  Talbot]{Marak_Cousty_Najman_Talbot2009}
L.~Marak, J.~Cousty, L.~Najman, and H.~Talbot.
\newblock 4d morphological segmentation and the miccai lv-segmentation grand
  challenge.
\newblock \url{http://hdl.handle.net/10380/3085}, 07 2009.

\bibitem[Marcus(1992)]{marcus-1992-overview}
Mitchell~P. Marcus.
\newblock Overview of the fifth {DARPA} speech and natural language workshop.
\newblock In \emph{Speech and Natural Language: Proceedings of a Workshop Held
  at Harriman, New York, {F}ebruary 23-26, 1992}, 1992.
\newblock URL \url{https://aclanthology.org/H92-1001}.

\bibitem[Matsui et~al.(1993)Matsui, Noumi, Yamashita, Wakahara, and
  Yoshimuro]{DBLP:conf/icdar/MatsuiNYWY93}
T.~Matsui, T.~Noumi, I.~Yamashita, T.~Wakahara, and M.~Yoshimuro.
\newblock State of the art of handwritten numeral recognition in japan-the
  results of the first {IPTP} character recognition competition.
\newblock In \emph{2nd International Conference Document Analysis and
  Recognition, {ICDAR} '93, October 20-22, 1993, Tsukuba City, Japan}, pages
  391--396. {IEEE} Computer Society, 1993.
\newblock \doi{10.1109/ICDAR.1993.395709}.
\newblock URL \url{https://doi.org/10.1109/ICDAR.1993.395709}.

\bibitem[Meyer and Saez-Rodriguez(2021)]{MEYER2021636}
Pablo Meyer and Julio Saez-Rodriguez.
\newblock Advances in systems biology modeling: 10 years of crowdsourcing dream
  challenges.
\newblock \emph{Cell Systems}, 12\penalty0 (6):\penalty0 636--653, 2021.
\newblock ISSN 2405-4712.
\newblock \doi{https://doi.org/10.1016/j.cels.2021.05.015}.
\newblock URL
  \url{https://www.sciencedirect.com/science/article/pii/S2405471221002015}.

\bibitem[Milani et~al.(2020)Milani, Topin, Houghton, Guss, Mohanty, Nakata,
  Vinyals, and Kuno]{pmlr-v123-milani20a}
Stephanie Milani, Nicholay Topin, Brandon Houghton, William~H. Guss, Sharada~P.
  Mohanty, Keisuke Nakata, Oriol Vinyals, and Noboru~Sean Kuno.
\newblock Retrospective analysis of the 2019 minerl competition on sample
  efficient reinforcement learning.
\newblock In Hugo~Jair Escalante and Raia Hadsell, editors, \emph{Proceedings
  of the NeurIPS 2019 Competition and Demonstration Track}, volume 123 of
  \emph{Proceedings of Machine Learning Research}, pages 203--214. PMLR, 08--14
  Dec 2020.
\newblock URL \url{https://proceedings.mlr.press/v123/milani20a.html}.

\bibitem[Palmero et~al.(2021)Palmero, Barquero, J{\'{u}}nior, Clap{\'{e}}s,
  N{\'{u}}{\~{n}}ez, Curto, Smeureanu, Selva, Zhang, Saeteros,
  Gallardo{-}Pujol, Guilera, Leiva, Han, Feng, He, Tu, Moeslund, Guyon, and
  Escalera]{DBLP:conf/iccv/PalmeroBJCNCSSZ21}
Cristina Palmero, Germ{\'{a}}n Barquero, J{\'{u}}lio C. S.~Jacques
  J{\'{u}}nior, Albert Clap{\'{e}}s, Johnny N{\'{u}}{\~{n}}ez, David Curto,
  Sorina Smeureanu, Javier Selva, Zejian Zhang, David Saeteros, David
  Gallardo{-}Pujol, Georgina Guilera, David Leiva, Feng Han, Xiaoxue Feng,
  Jennifer He, Wei{-}Wei Tu, Thomas~B. Moeslund, Isabelle Guyon, and Sergio
  Escalera.
\newblock Chalearn {LAP} challenges on self-reported personality recognition
  and non-verbal behavior forecasting during social dyadic interactions:
  Dataset, design, and results.
\newblock In Cristina Palmero, J{\'{u}}lio C. S.~Jacques J{\'{u}}nior, Albert
  Clap{\'{e}}s, Isabelle Guyon, Wei{-}Wei Tu, Thomas~B. Moeslund, and Sergio
  Escalera, editors, \emph{ChaLearn {LAP} Challenge on Understanding Social
  Behavior in Dyadic and Small Group Interactions, {DYAD} 2021, held in
  conjunction with {ICCV} 2021, Virtual, October 16, 2021}, volume 173 of
  \emph{Proceedings of Machine Learning Research}, pages 4--52. {PMLR}, 2021.
\newblock URL \url{https://proceedings.mlr.press/v173/palmero22b.html}.

\bibitem[Pavao et~al.(2023)Pavao, Guyon, Letournel, Tran, Baro, Escalante,
  Escalera, Thomas, and Xu]{codalab_competitions_JMLR}
Adrien Pavao, Isabelle Guyon, Anne-Catherine Letournel, Dinh-Tuan Tran, Xavier
  Baro, Hugo~Jair Escalante, Sergio Escalera, Tyler Thomas, and Zhen Xu.
\newblock Codalab competitions: An open source platform to organize scientific
  challenges.
\newblock \emph{Journal of Machine Learning Research}, 24\penalty0
  (198):\penalty0 1--6, 2023.
\newblock URL \url{http://jmlr.org/papers/v24/21-1436.html}.

\bibitem[Ramachandran et~al.(2021)Ramachandran, Sistu, McDonald, and
  Yogamani]{Fisheye_AD}
Saravanabalagi Ramachandran, Ganesh Sistu, John~B. McDonald, and Senthil~Kumar
  Yogamani.
\newblock Woodscape fisheye semantic segmentation for autonomous driving -
  {CVPR} 2021 omnicv workshop challenge.
\newblock \emph{CoRR}, abs/2107.08246, 2021.
\newblock URL \url{https://arxiv.org/abs/2107.08246}.

\bibitem[Redmon et~al.(2016)Redmon, Divvala, Girshick, and
  Farhadi]{DBLP:conf/cvpr/RedmonDGF16}
Joseph Redmon, Santosh~Kumar Divvala, Ross~B. Girshick, and Ali Farhadi.
\newblock You only look once: Unified, real-time object detection.
\newblock In \emph{2016 {IEEE} Conference on Computer Vision and Pattern
  Recognition, {CVPR} 2016, Las Vegas, NV, USA, June 27-30, 2016}, pages
  779--788. {IEEE} Computer Society, 2016.
\newblock \doi{10.1109/CVPR.2016.91}.
\newblock URL \url{https://doi.org/10.1109/CVPR.2016.91}.

\bibitem[Rowe et~al.(2010)Rowe, Wood, Link, and Simon]{OVER2001369}
Brent~R. Rowe, Dallas~W. Wood, Albert~N. Link, and Diglio~A. Simon.
\newblock Economic impact assessment of nist’s text retrieval conference
  (trec) program.
\newblock NIST Final Report, 2010.

\bibitem[Russakovsky et~al.(2015)Russakovsky, Deng, Su, Krause, Satheesh, Ma,
  Huang, Karpathy, Khosla, Bernstein, Berg, and Fei-Fei]{Russakovsky2015}
Olga Russakovsky, Jia Deng, Hao Su, Jonathan Krause, Sanjeev Satheesh, Sean Ma,
  Zhiheng Huang, Andrej Karpathy, Aditya Khosla, Michael Bernstein,
  Alexander~C. Berg, and Li~Fei-Fei.
\newblock Imagenet large scale visual recognition challenge.
\newblock \emph{International Journal of Computer Vision}, 115\penalty0
  (3):\penalty0 211--252, Dec 2015.
\newblock ISSN 1573-1405.
\newblock \doi{10.1007/s11263-015-0816-y}.
\newblock URL \url{https://doi.org/10.1007/s11263-015-0816-y}.

\bibitem[Scully et~al.(2008)Scully, Magnotta, Gasparovic, Pelligrimo, Feis, and
  Bockholt]{scully}
M.~Scully, V.~Magnotta, C.~Gasparovic, P.~Pelligrimo, D.~Feis, and H.~Bockholt.
\newblock 3d segmentation in the clinic: A grand challenge ii at miccai 2008 -
  ms lesion segmentation.
\newblock \url{http://hdl.handle.net/10380/1449}, 07 2008.

\bibitem[Shah et~al.(2022)Shah, Wang, Wild, Milani, Kanervisto, Goecks,
  Waytowich, Watkins-Valls, Prakash, Mills, Garg, Fries, Souly, Shern, del
  Castillo, and Lieberum]{https://doi.org/10.48550/arxiv.2204.07123}
Rohin Shah, Steven~H. Wang, Cody Wild, Stephanie Milani, Anssi Kanervisto,
  Vinicius~G. Goecks, Nicholas Waytowich, David Watkins-Valls, Bharat Prakash,
  Edmund Mills, Divyansh Garg, Alexander Fries, Alexandra Souly, Chan~Jun
  Shern, Daniel del Castillo, and Tom Lieberum.
\newblock Retrospective on the 2021 basalt competition on learning from human
  feedback, 2022.
\newblock URL \url{https://arxiv.org/abs/2204.07123}.

\bibitem[Simonyan and Zisserman(2015)]{DBLP:journals/corr/SimonyanZ14a}
Karen Simonyan and Andrew Zisserman.
\newblock Very deep convolutional networks for large-scale image recognition.
\newblock In Yoshua Bengio and Yann LeCun, editors, \emph{3rd International
  Conference on Learning Representations, {ICLR} 2015, San Diego, CA, USA, May
  7-9, 2015, Conference Track Proceedings}, 2015.
\newblock URL \url{http://arxiv.org/abs/1409.1556}.

\bibitem[Sincan et~al.(2021)Sincan, J{\'{u}}nior, Escalera, and
  Keles]{DBLP:conf/cvpr/SincanJEK21}
Ozge~Mercanoglu Sincan, J{\'{u}}lio C. S.~Jacques J{\'{u}}nior, Sergio
  Escalera, and Hacer~Yalim Keles.
\newblock Chalearn {LAP} large scale signer independent isolated sign language
  recognition challenge: Design, results and future research.
\newblock In \emph{{IEEE} Conference on Computer Vision and Pattern Recognition
  Workshops, {CVPR} Workshops 2021, virtual, June 19-25, 2021}, pages
  3472--3481. Computer Vision Foundation / {IEEE}, 2021.
\newblock \doi{10.1109/CVPRW53098.2021.00386}.
\newblock URL
  \url{https://openaccess.thecvf.com/content/CVPR2021W/ChaLearn/html/Sincan\_ChaLearn\_LAP\_Large\_Scale\_Signer\_Independent\_Isolated\_Sign\_Language\_Recognition\_CVPRW\_2021\_paper.html}.

\bibitem[Stefanini et~al.(2021)Stefanini, Cornia, Baraldi, Cascianelli,
  Fiameni, and Cucchiara]{DBLP:journals/corr/abs-2107-06912}
Matteo Stefanini, Marcella Cornia, Lorenzo Baraldi, Silvia Cascianelli,
  Giuseppe Fiameni, and Rita Cucchiara.
\newblock From show to tell: {A} survey on image captioning.
\newblock \emph{CoRR}, abs/2107.06912, 2021.
\newblock URL \url{https://arxiv.org/abs/2107.06912}.

\bibitem[Stolovitzky et~al.(2009)Stolovitzky, Prill, and
  Califano]{https://doi.org/10.1111/j.1749-6632.2009.04497.x}
Gustavo Stolovitzky, Robert~J. Prill, and Andrea Califano.
\newblock Lessons from the dream2 challenges.
\newblock \emph{Annals of the New York Academy of Sciences}, 1158\penalty0
  (1):\penalty0 159--195, 2009.
\newblock \doi{https://doi.org/10.1111/j.1749-6632.2009.04497.x}.
\newblock URL
  \url{https://nyaspubs.onlinelibrary.wiley.com/doi/abs/10.1111/j.1749-6632.2009.04497.x}.

\bibitem[Sun et~al.(2022)Sun, Nguyen, Allaway, Wang, Chung, Yu, Mason,
  Dimitrovsky, Ericson, Li, Guan, Israel, Olar, Pataki, Stolovitzky, Guinney,
  Gulko, Frazier, Chen, Costello, Bridges, and
  Community]{10.1001/jamanetworkopen.2022.27423}
Dongmei Sun, Thanh~M. Nguyen, Robert~J. Allaway, Jelai Wang, Verena Chung,
  Thomas~V. Yu, Michael Mason, Isaac Dimitrovsky, Lars Ericson, Hongyang Li,
  Yuanfang Guan, Ariel Israel, Alex Olar, Balint~Armin Pataki, Gustavo
  Stolovitzky, Justin Guinney, Percio~S. Gulko, Mason~B. Frazier, Jake~Y. Chen,
  James~C. Costello, Jr~Bridges, S.~Louis, and RA2-DREAM~Challenge Community.
\newblock {A Crowdsourcing Approach to Develop Machine Learning Models to
  Quantify Radiographic Joint Damage in Rheumatoid Arthritis}.
\newblock \emph{JAMA Network Open}, 5\penalty0 (8):\penalty0
  e2227423--e2227423, 08 2022.
\newblock ISSN 2574-3805.
\newblock \doi{10.1001/jamanetworkopen.2022.27423}.
\newblock URL \url{https://doi.org/10.1001/jamanetworkopen.2022.27423}.

\bibitem[Szegedy et~al.(2015)Szegedy, Liu, Jia, Sermanet, Reed, Anguelov,
  Erhan, Vanhoucke, and Rabinovich]{DBLP:conf/cvpr/SzegedyLJSRAEVR15}
Christian Szegedy, Wei Liu, Yangqing Jia, Pierre Sermanet, Scott~E. Reed,
  Dragomir Anguelov, Dumitru Erhan, Vincent Vanhoucke, and Andrew Rabinovich.
\newblock Going deeper with convolutions.
\newblock In \emph{{IEEE} Conference on Computer Vision and Pattern
  Recognition, {CVPR} 2015, Boston, MA, USA, June 7-12, 2015}, pages 1--9.
  {IEEE} Computer Society, 2015.
\newblock \doi{10.1109/CVPR.2015.7298594}.
\newblock URL \url{https://doi.org/10.1109/CVPR.2015.7298594}.

\bibitem[Tarca et~al.(2020)Tarca, Pataki, Romero, Sirota, Guan, Kutum,
  Gomez-Lopez, Done, Bhatti, Yu, Andreoletti, Chaiworapongsa, Hassan, Hsu,
  Aghaeepour, Stolovitzky, Csabai, and Costello]{Tarca2020.06.05.130971}
Adi~L. Tarca, B{\'a}lint~{\'A}rmin Pataki, Roberto Romero, Marina Sirota,
  Yuanfang Guan, Rintu Kutum, Nardhy Gomez-Lopez, Bogdan Done, Gaurav Bhatti,
  Thomas Yu, Gaia Andreoletti, Tinnakorn Chaiworapongsa, Sonia~S. Hassan,
  Chaur-Dong Hsu, Nima Aghaeepour, Gustavo Stolovitzky, Istvan Csabai, and
  James~C. Costello.
\newblock Crowdsourcing assessment of maternal blood multi-omics for predicting
  gestational age and preterm birth.
\newblock \emph{bioRxiv}, 2020.
\newblock \doi{10.1101/2020.06.05.130971}.
\newblock URL
  \url{https://www.biorxiv.org/content/early/2020/06/06/2020.06.05.130971}.

\bibitem[Timofte et~al.(2017)Timofte, Agustsson, Gool, Yang, Zhang, Lim, Son,
  Kim, Nah, Lee, Wang, Tian, Yu, Zhang, Wu, Dong, Lin, Qiao, Loy, Bae, Yoo,
  Han, Ye, Choi, Kim, Fan, Yu, Han, Liu, Yu, Wang, Shi, Wang, Huang, Chen,
  Zhang, Zuo, Tang, Luo, Li, Fu, Cao, Heng, Bui, Le, Duan, Tao, Wang, Lin,
  Pang, Xu, Zhao, Xu, Pan, Sun, Zhang, Song, Dai, Qin, Huynh, Guo, Mousavi, Vu,
  Monga, Cruz, Egiazarian, Katkovnik, Mehta, Jain, Agarwalla, Praveen, Zhou,
  Wen, Zhu, Xia, Wang, and Guo]{DBLP:conf/cvpr/TimofteAG0ZLSKN17}
Radu Timofte, Eirikur Agustsson, Luc~Van Gool, Ming{-}Hsuan Yang, Lei Zhang,
  Bee Lim, Sanghyun Son, Heewon Kim, Seungjun Nah, Kyoung~Mu Lee, Xintao Wang,
  Yapeng Tian, Ke~Yu, Yulun Zhang, Shixiang Wu, Chao Dong, Liang Lin, Yu~Qiao,
  Chen~Change Loy, Woong Bae, Jae~Jun Yoo, Yoseob Han, Jong~Chul Ye, Jae{-}Seok
  Choi, Munchurl Kim, Yuchen Fan, Jiahui Yu, Wei Han, Ding Liu, Haichao Yu,
  Zhangyang Wang, Honghui Shi, Xinchao Wang, Thomas~S. Huang, Yunjin Chen, Kai
  Zhang, Wangmeng Zuo, Zhimin Tang, Linkai Luo, Shaohui Li, Min Fu, Lei Cao,
  Wen Heng, Giang Bui, Truc Le, Ye~Duan, Dacheng Tao, Ruxin Wang, Xu~Lin,
  Jianxin Pang, Jinchang Xu, Yu~Zhao, Xiangyu Xu, Jin{-}shan Pan, Deqing Sun,
  Yujin Zhang, Xibin Song, Yuchao Dai, Xueying Qin, Xuan{-}Phung Huynh,
  Tiantong Guo, Hojjat~Seyed Mousavi, Tiep~Huu Vu, Vishal Monga,
  Crist{\'{o}}v{\~{a}}o Cruz, Karen~O. Egiazarian, Vladimir Katkovnik, Rakesh
  Mehta, Arnav~Kumar Jain, Abhinav Agarwalla, Ch~V.~Sai Praveen, Ruofan Zhou,
  Hongdiao Wen, Che Zhu, Zhiqiang Xia, Zhengtao Wang, and Qi~Guo.
\newblock {NTIRE} 2017 challenge on single image super-resolution: Methods and
  results.
\newblock In \emph{2017 {IEEE} Conference on Computer Vision and Pattern
  Recognition Workshops, {CVPR} Workshops 2017, Honolulu, HI, USA, July 21-26,
  2017}, pages 1110--1121. {IEEE} Computer Society, 2017.
\newblock \doi{10.1109/CVPRW.2017.149}.
\newblock URL \url{https://doi.org/10.1109/CVPRW.2017.149}.

\bibitem[Turner et~al.(2021)Turner, Eriksson, McCourt, Kiili, Laaksonen, Xu,
  and Guyon]{pmlr-v133-turner21a}
Ryan Turner, David Eriksson, Michael McCourt, Juha Kiili, Eero Laaksonen, Zhen
  Xu, and Isabelle Guyon.
\newblock Bayesian optimization is superior to random search for machine
  learning hyperparameter tuning: Analysis of the black-box optimization
  challenge 2020.
\newblock In Hugo~Jair Escalante and Katja Hofmann, editors, \emph{Proceedings
  of the NeurIPS 2020 Competition and Demonstration Track}, volume 133 of
  \emph{Proceedings of Machine Learning Research}, pages 3--26. PMLR, 06--12
  Dec 2021.
\newblock URL \url{https://proceedings.mlr.press/v133/turner21a.html}.

\bibitem[Urmson et~al.(2007)Urmson, Anhalt, Bagnell, Baker, Bittner, Dolan,
  Duggins, Ferguson, Galatali, Geyer, Gittleman, Harbaugh, Hebert, Howard,
  Kelly, Kohanbash, Likhachev, Miller, Peterson, Rajkumar, Rybski, Salesky,
  Scherer, Seo, Simmons, Singh, Snider, Stentz, Whittaker, and
  Ziglar]{DarpaTartanTeam}
Christopher Urmson, Joshua Anhalt, J.~Andrew~(Drew) Bagnell, Christopher~R.
  Baker, Robert~E. Bittner, John~M. Dolan, David Duggins, David Ferguson,
  Tugrul Galatali, Hartmut Geyer, Michele Gittleman, Sam Harbaugh, Martial
  Hebert, Thomas Howard, Alonzo Kelly, David Kohanbash, Maxim Likhachev, Nick
  Miller, Kevin Peterson, Raj Rajkumar, Paul Rybski, Bryan Salesky, Sebastian
  Scherer, Young-Woo Seo, Reid Simmons, Sanjiv Singh, Jarrod~M. Snider,
  Anthony~(Tony) Stentz, William (Red)~L. Whittaker, and Jason Ziglar.
\newblock Tartan racing: A multi-modal approach to the darpa urban challenge.
\newblock Technical report, Carnegie Mellon University, Pittsburgh, PA, April
  2007.

\bibitem[Visser(2016)]{DBLP:journals/ki/Visser16a}
Ubbo Visser.
\newblock 20 years of robocup.
\newblock \emph{K{\"{u}}nstliche Intell.}, 30\penalty0 (3-4):\penalty0
  217--220, 2016.
\newblock \doi{10.1007/s13218-016-0439-7}.
\newblock URL \url{https://doi.org/10.1007/s13218-016-0439-7}.

\bibitem[Voorhees(2001)]{DBLP:journals/nle/Voorhees01}
Ellen~M. Voorhees.
\newblock The {TREC} question answering track.
\newblock \emph{Nat. Lang. Eng.}, 7\penalty0 (4):\penalty0 361--378, 2001.
\newblock \doi{10.1017/S1351324901002789}.
\newblock URL \url{https://doi.org/10.1017/S1351324901002789}.

\bibitem[Voorhees and Harman(1998)]{DBLP:conf/tipster/VoorheesH98}
Ellen~M. Voorhees and Donna Harman.
\newblock The text retrieval conferences {(TRECS)}.
\newblock In \emph{{TIPSTER} {TEXT} {PROGRAM} {PHASE} {III:} Proceedings of a
  Workshop held at Baltimore, MD, USA, October 13-15, 1998}, pages 241--273.
  Morgan Kaufmann, 1998.
\newblock \doi{10.3115/1119089.1119127}.
\newblock URL \url{https://aclanthology.org/X98-1031/}.

\bibitem[Voorhees and Harman(2005)]{DBLP:journals/ipm/Muresan07}
Ellen~M. Voorhees and Donna~K. Harman, editors.
\newblock \emph{{TREC} - Experiment and Evaluation in Information Retrieval}.
\newblock The {MIT} Press, Cambridge, 2005.

\bibitem[Wan et~al.(2017)Wan, Escalera, Anbarjafari, Escalante, Bar{\'{o}},
  Guyon, Madadi, Allik, Gorbova, Lin, and Xie]{DBLP:conf/iccvw/WanEAEBGMAGLX17}
Jun Wan, Sergio Escalera, Gholamreza Anbarjafari, Hugo~Jair Escalante, Xavier
  Bar{\'{o}}, Isabelle Guyon, Meysam Madadi, Juri Allik, Jelena Gorbova, Chi
  Lin, and Yiliang Xie.
\newblock Results and analysis of chalearn {LAP} multi-modal isolated and
  continuous gesture recognition, and real versus fake expressed emotions
  challenges.
\newblock In \emph{2017 {IEEE} International Conference on Computer Vision
  Workshops, {ICCV} Workshops 2017, Venice, Italy, October 22-29, 2017}, pages
  3189--3197. {IEEE} Computer Society, 2017.
\newblock \doi{10.1109/ICCVW.2017.377}.
\newblock URL \url{https://doi.org/10.1109/ICCVW.2017.377}.

\bibitem[Wan et~al.(2020)Wan, Guo, Escalera, Escalante, and
  Li]{DBLP:series/synthesis/2020Wan}
Jun Wan, Guodong Guo, Sergio Escalera, Hugo~Jair Escalante, and Stan~Z. Li.
\newblock \emph{Multi-Modal Face Presentation Attack Detection}.
\newblock Synthesis Lectures on Computer Vision. Morgan {\&} Claypool
  Publishers, 2020.
\newblock \doi{10.2200/S01032ED1V01Y202007COV017}.
\newblock URL \url{https://doi.org/10.2200/S01032ED1V01Y202007COV017}.

\bibitem[Wang et~al.(2021{\natexlab{a}})Wang, Hwang, Wang, Liu, Kim, Hsu, Cai,
  Zhang, Jiang, and Gu]{wang2021rod2021}
Yizhou Wang, Jenq-Neng Hwang, Gaoang Wang, Hui Liu, Kwang-Ju Kim, Hung-Min Hsu,
  Jiarui Cai, Haotian Zhang, Zhongyu Jiang, and Renshu Gu.
\newblock Rod2021 challenge: A summary for radar object detection challenge for
  autonomous driving applications.
\newblock In \emph{Proceedings of the 2021 International Conference on
  Multimedia Retrieval}, pages 553--559, 2021{\natexlab{a}}.

\bibitem[Wang et~al.(2021{\natexlab{b}})Wang, Jiang, Li, Hwang, Xing, and
  Liu]{9353210}
Yizhou Wang, Zhongyu Jiang, Yudong Li, Jenq-Neng Hwang, Guanbin Xing, and Hui
  Liu.
\newblock Rodnet: A real-time radar object detection network cross-supervised
  by camera-radar fused object 3d localization.
\newblock \emph{IEEE Journal of Selected Topics in Signal Processing},
  15\penalty0 (4):\penalty0 954--967, 2021{\natexlab{b}}.
\newblock \doi{10.1109/JSTSP.2021.3058895}.

\bibitem[Yadav and Bottou(2019)]{qmnist-2019}
Chhavi Yadav and L\'{e}on Bottou.
\newblock Cold case: The lost mnist digits.
\newblock In \emph{Advances in Neural Information Processing Systems 32}.
  Curran Associates, Inc., 2019.

\end{thebibliography}

\end{document}